\documentclass[10pt,journal]{IEEEtran}

\usepackage{outline}
\usepackage{pmgraph}
\usepackage[normalem]{ulem}
\usepackage[utf8]{inputenc}
\usepackage{amssymb}
\usepackage{hyperref}
\usepackage{amsmath}
\usepackage{graphicx}
\usepackage{times}
\usepackage{xcolor}
\usepackage{xspace}
\usepackage[colorinlistoftodos]{todonotes} 
\usepackage{cite}
\usepackage{bm} 
\usepackage{url}

\usepackage{epstopdf}
\AppendGraphicsExtensions{.tiff}
\graphicspath{{fig/}} 

\usepackage{epsfig}
\usepackage{tikz}
\usetikzlibrary{spy}
\usepackage{algpseudocode}
\usepackage{algorithm}
\usepackage{mathrsfs}

\usepackage{nicefrac}       
\usepackage{booktabs}       

\usepackage{amsmath,graphicx,caption,mathtools,amssymb,amsthm}
\usepackage{graphicx}
\usepackage{multirow}
\usepackage{subcaption}
\usepackage{accents}

\newcommand{\add}[1] {\textcolor{black}{#1}} 

\long\def\comment#1{} 

\newcommand{\xmath}[1] {\ensuremath{#1}\xspace}
\newcommand{\blmath}[1] {\xmath{\bm{#1}}}

\newcommand{\B}{\blmath{B}}

\newcommand{\E}{\blmath{E}}

\newcommand{\x}{\blmath{x}}

\newcommand{\z}{\blmath{z}}

\newcommand{\Bb}{{\blmath B}}

\newcommand{\Db}{{\blmath D}}
\newcommand{\Eb}{{\blmath E}}

\newcommand{\Ib}{{\blmath I}}

\newcommand{\Mb}{{\blmath M}}

\newcommand{\Sb}{{\blmath S}}

\newcommand{\bb}{{\blmath b}}

\newcommand{\kb}{{\blmath k}}

\newcommand{\nb}{{\blmath n}}

\newcommand{\rb}{{\blmath r}}

\newcommand{\vb}{{\blmath v}}
\newcommand{\wb}{{\blmath w}}
\newcommand{\xb}{{\blmath x}}
\newcommand{\yb}{{\blmath y}}
\newcommand{\zb}{{\blmath z}}

\newcommand{\Bc}{\mathcal{B}}

\newcommand{\Tc}{\mathcal{T}}

\newcommand{\Phib}{{\boldsymbol {\Phi}}}
\newcommand{\Psib}{{\boldsymbol {\Psi}}}
\newcommand{\Thetab}{{\boldsymbol {\Theta}}}

\newcommand{\Rd}{{\mathbb R}}

\newcommand{\psib}{{\boldsymbol{\psi}}}

\newcommand{\chib}{{\boldsymbol {\chi}}}
\newcommand{\xib}{{\boldsymbol {\xi}}}

\newcommand{\Pc}{{{\mathcal P}}}
\newcommand{\Ec}{{{\mathcal E}}}
\newcommand{\Dc}{{{\mathcal D}}}

\newcommand{\beq}{\begin{equation}}
\newcommand{\eeq}{\end{equation}}
\newcommand{\beqa}{\begin{eqnarray}}
\newcommand{\eeqa}{\end{eqnarray}}
\newcommand{\Fc}{{\mathcal F}}

\newcommand{\Lambdab}{\boldsymbol{\Lambda}}

\begin{document}

\title{Geometric Approaches to Increase the Expressivity of  Deep Neural Networks for MR Reconstruction}

\author{Eunju~Cha, Gyutaek Oh, and~Jong~Chul~Ye$^{*}$,~\IEEEmembership{Fellow,~IEEE}
\thanks{E. Cha, G. Oh, and J.C. Ye are with the Department of Bio and Brain Engineering, Korea Advanced Institute of Science and Technology (KAIST), 
		Daejeon 34141, Republic of Korea (e-mail: \{eunju.cha, okt0711, jong.ye\}@kaist.ac.kr).  
		
		This work is supported by National Research Foundation of Korea,
Grant number NRF2016R1A2B3008104.}
}		

\maketitle

\begin{abstract}
Recently, deep learning approaches have been extensively investigated to reconstruct images from accelerated magnetic resonance image (MRI) acquisition. Although these approaches provide significant performance gain compared to compressed sensing MRI (CS-MRI), it is not clear how to choose a suitable network architecture to balance the trade-off between network complexity and performance. 
\add{Recently, it was shown that an encoder-decoder convolutional neural network (CNN) can be interpreted as a piecewise linear basis-like representation, whose specific representation is determined by the ReLU activation patterns for a given input image. Thus, the expressivity
or the representation power is determined by the number of piecewise linear regions.}
As an extension of this geometric understanding, this paper proposes a systematic geometric approach using bootstrapping and subnetwork aggregation using an attention module to increase the expressivity of the underlying neural network. Our method can be implemented in both k-space domain and image domain that can be trained in an end-to-end manner. Experimental results show that the proposed schemes significantly improve reconstruction performance with negligible complexity increases.
\end{abstract}


\section{Introduction}
\label{Intro}

\IEEEPARstart{M}{agnetic} resonance imaging (MRI) is a valuable imaging method for diagnosis. However, its long scan time still remains a challenge
for MRI. To address this issue, researchers have investigated various MR acceleration techniques,  where image reconstruction is performed
from sub-sampled $k$-space measurements.
Specifically,
for a given under-sampling pattern $\Lambda$,
 the  $k$-space measurement data  for accelerated MR  is  given by
 \begin{eqnarray}\label{eq:fwd}
\hat\yb & :=\Pc_\Lambda[\hat \xb] 
\end{eqnarray}
where
  the downsampling operator $\Pc_\Lambda$ 
is defined as
  \begin{eqnarray}
  \left[\Pc_\Lambda[\hat \xb] \right]_i= \begin{cases}  \widehat x[i], &i \in \Lambda \\
0, &  \mbox{otherwise} \end{cases}   \   .
  \end{eqnarray}
and the spatial Fourier transform of the unknown image $x:\Rd^2\to\Rd$ is 
\begin{equation}
\hat{x}(\kb)=\mathcal{F}[x](\kb):=\int_{\Rd^d} e^{-\add{j}\kb\cdot \rb}x(\rb)d\rb,
\end{equation}
with spatial frequency $\kb\in\Rd^2$ and $\add{j}=\sqrt{-1}$. 
 Then, the image reconstruction problem for accelerated MRI  is to estimate
the unknown $x(\rb)$ (or its discretized version $\xb$) from the subsampled measurement $\hat\yb$.

To address the ill-posedness of the inverse problem from the undersampled $k$-space measurements,
many algorithms have been developed  by exploiting additional
prior information, 
such as multi-coil redundancy as in 
parallel MRI (pMRI) \cite{griswold2002generalized, liang2009accelerating}, or sparsity in the case of compressed sensing MRI (CS-MRI) \cite{lustig2008compressed,jung2009k, shin2014calibrationless, jin2015general}. 
Recently, 
deep learning approaches have been extensively explored as  promising alternatives  for accelerated MRI thanks to its high performance in spite of significantly reduced run-time complexity \cite{wang2016accelerating, kwon2017parallel, hammernik2018learning, lee2017deep, jin2017deep, lee2018deep, zhu2018image,sun2018compressed, akccakaya2019scan, han2019k,lee2019k}. 
For example,  Kwon et al \cite{kwon2017parallel} proposed multilayer perceptron for parallel MRI.
Wang et al \cite{wang2016accelerating} 
 employed a deep learning prior as an initialization or regularization term for compressed sensing reconstruction. 
 Hammernik et al \cite{hammernik2018learning} proposed a variational network  by learning each step of
unrolled compressed sensing iterations. Learning-based alternating directional methods of multiplier (ADMM) was also 
proposed by unrolling ADMM steps \cite{sun2016deep}.
Since the introduction of these pioneering works, this area has been populated with many innovative and interesting
network architectures such as image domain approaches\cite{lee2018deep}, 
$k$-space domain approaches \cite{akccakaya2019scan,han2019k,lee2019k}, hybrid domain approaches \cite{eo2018kiki},  domain transform learning
approach \cite{zhu2018image}, \add{recursive learning approach \cite{sun2018compressed}}, etc.

Specifically, the basic idea of deep learning for compressed sensing MRI is to design a neural network $\Tc_\Thetab$ parameterized with trainable parameters $\Thetab$ by minimizing an 
 empirical loss:
\begin{eqnarray}
\min_\Thetab \frac{1}{T}\sum_{t=1}^T \ell\left(\vb^{(t)},\Tc_\Thetab\left(\zb^{(t)}\right)\right)
\end{eqnarray}
where  $\ell$ is a performance metric such as $l_2$ norm,
and $\{\vb^{(t)},\zb^{(t)}\}_{t=1}^T$ refers to $T$ training 
data,
defined as 
\begin{align}
(\vb,\zb) &:= \begin{cases} (\hat\xb , \hat\yb),& \mbox{($k$-space learning)}\\
\left( \Fc^{-1}(\hat \xb),\Fc^{-1}(\hat\yb)\right), & \mbox{(image learning)} \end{cases}
\end{align}

Aside from using different loss functions such as GAN loss \cite{goodfellow2014generative} or perceptual loss \cite{johnson2016perceptual}
as in \add{\cite{yang2017dagan,mardani2017deep,quan2018compressed}},
one of the important research efforts in this field is to develop novel network architecture that provides high quality
image reconstruction. Unfortunately,
many of these new network architectures are usually associated with increased network complexity,
so it is not clear whether the performance increase is due to the unique network topology or
the increased number of network parameters.
\add{Therefore, it is important to analyze the ``expressivity'' or ``representation power'' of a neutral network, which indicates its ability to approximate the desired function. More specifically, enhancing the expressivity of neural network is directly related to the improvement in reconstruction performance thanks to its elevated flexibility in approximating a nonlinear mapping.}

Recently, inspired by the so-called convolutional framelet interpretation of CNN \cite{ye2018deep},
our group showed that
encoder-decoder CNN can be understood as a piecewise linear framelet representation whose
basis are adaptively chosen by the rectified linear unit (ReLU) activation patterns determined by the network input \cite{ye2019understanding}.
\add{We further showed that the increased network width (i.e. number of filter channels) increases the number of piecewise linear regions in which one of them is selected depending on the input data. Therefore, this increase directly leads to an improvement in the expressivity of the neural network.}
Inspired by these geometrical understanding of the encoder-decoder CNN \cite{ye2019understanding},
here we provide  additional geometric insights of CNNs that
lead to  a systematic approach to further improve the expressivity of the neural network.
Specially, we show that a novel
attention scheme combined with  bootstrapping and subnetwork aggregation  improves network expressivity with
minimal complexity increase.
In particular,  
the attention module is shown to provide a redundant
 representation and increased number of piecewise linear regions that improve the expressivity of the network.
Thanks to the increased expressivity,
the proposed network modification  improves the reconstruction performance. 

As a proof of concept, 
we provide several modifications of the popular neural network baseline  U-Net \cite{ronneberger2015u} that is often used
for MR reconstruction from sparse samples \cite{jin2017deep,lee2017deep,han2017deep}.
  Experimental results show that the modified U-Net produces significantly better reconstruction results with negligible complexity increases.

\begin{figure}[!bt] 	
\center{ 
\includegraphics[width=9.0cm]{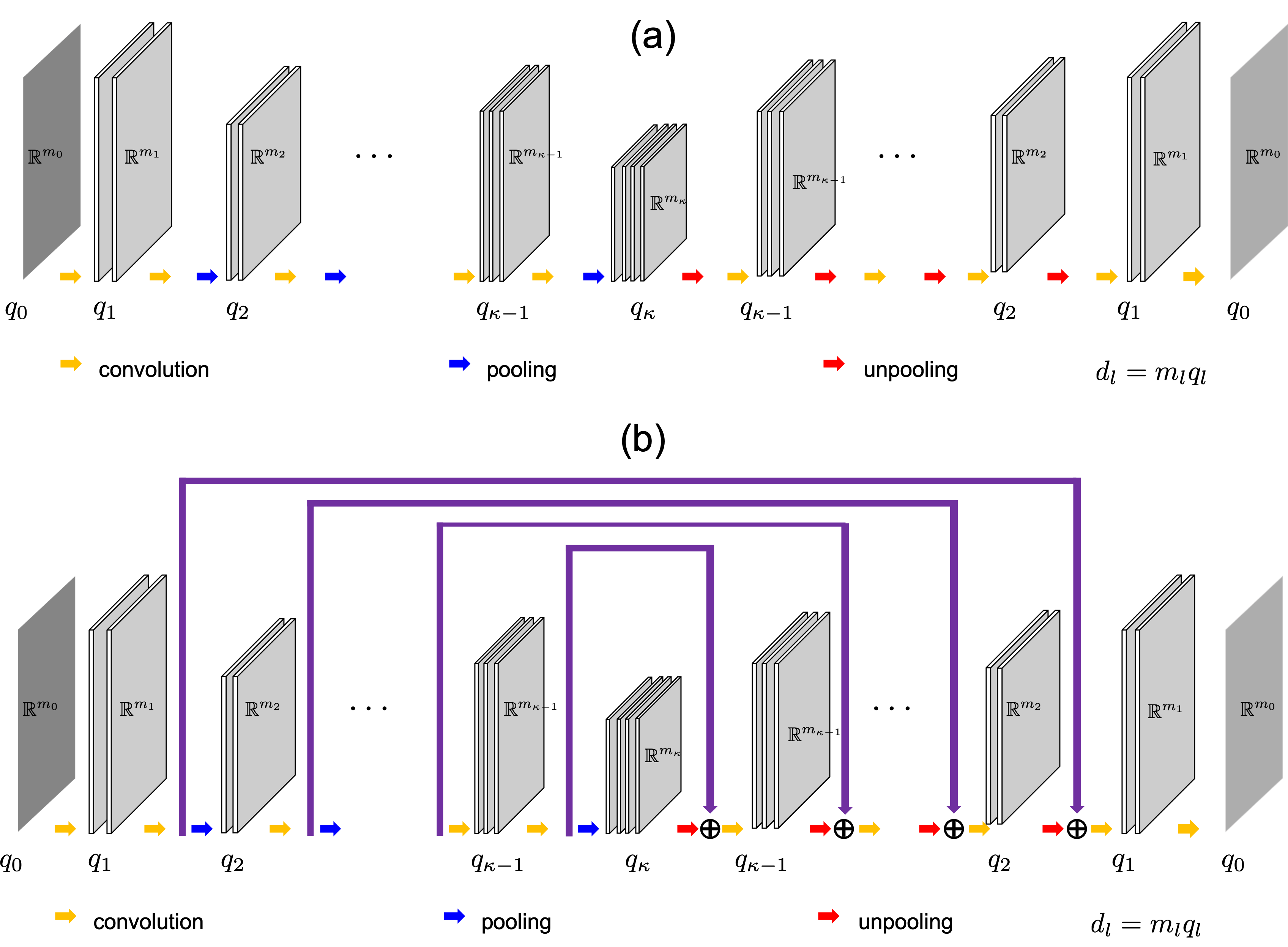}
}
\caption{Encoder-decoder CNNs (a) without skipped connection and (b) with skipped connection.}
\label{fig:linearCNN}
\end{figure}

\section{Geometry of  CNN: A Bird-Eye View}

To understand the theoretical background for the proposed method,
here we  provide a review of the recent geometric understanding on CNN \cite{ye2019understanding}.

\subsection{Definition}



Consider  encoder-decoder networks in Fig.~\ref{fig:linearCNN}.
For simplicity, here we consider a symmetric {configuration} so that
both encoder and decoder have the same number of layers, say $\kappa$;
the input and output dimensions for the encoder layer $\Ec^l$ and the decoder layer $\Dc^l$ are symmetric:
\begin{eqnarray}
\Ec^l:\Rd^{d_{l-1}} \mapsto \Rd^{d_l},  \quad 
\Dc^l:\Rd^{d_{l}} \mapsto \Rd^{d_{l-1}}, \quad l\in[\kappa]
\end{eqnarray}
where $[n]$ denotes the set $\{1,\cdots, n\}$.
 At the $l$-th layer, $m_l$ and $q_l$ denote the dimension of the signal, and  the number of filter channel, respectively. The length of filter
is assumed to be $r$.

We now define the $l$-th layer input signal for the encoder layer from $q_{l-1}$ number of input channels
\begin{equation}
\zb^{l-1}:=\begin{bmatrix} \zb_1^{l-1\top} & \cdots & \zb^{l-1\top}_{q_{l-1}} \end{bmatrix}^\top \in   \Rd^{d_{l-1}}, \quad
\end{equation}
where  $^\top$ denotes the transpose,
and $\zb_j^{l-1} \in \Rd^{m_{l-1}}$ refers to the $j$-th channel input with the dimension $m_{l-1}$.
The $l$-th layer output signal $\zb^l$ is similarly defined.
Then, 
we have the following representations of the convolution and pooling operation at the $l$-th encoder layer \cite{ye2019understanding}:
\begin{eqnarray}\label{eq:Enc}
 \z^l
 = \sigma\left(\Eb^{l\top} \z^{l-1}\right)
\end{eqnarray}
where $\sigma(\cdot)$ is defined as an element-by-element ReLU operation $\sigma(x)=\max\{x,0\}$, and 
   \begin{eqnarray}\label{eq:El}
\E^l= \begin{bmatrix} 
\Phib^l\circledast \psib^l_{1,1} & \cdots &  \Phib^l\circledast \psib^l_{q_l,1}  \\
  \vdots & \ddots & \vdots \\
\Phib^l\circledast \psib^l_{1,q_{l-1}} & \cdots &\Phib^l\circledast \psib^l_{q_{l},q_{l-1}}
 \end{bmatrix}
 \end{eqnarray}
 where $\Phib^l$ denotes the $m_l\times m_l$  matrix that represents the pooling operation at the $l$-th layer, and $\psib_{i,j}^l\in\Rd^r$ represents
the $l$-th layer encoder filter to generate the $i$-th channel output from the contribution of the $j$-th channel input,
and
 $\circledast$ represents a multi-channel convolution \cite{ye2019understanding}.
Note that the inclusion of the bias in \eqref{eq:Enc} can be readily done by including an additional row into $\Eb^l$ as a bias
and augmenting the last element of $\z^{l-1}$ as 1. Accordingly, without loss of generality, we assume that
\eqref{eq:Enc} is the general expression including the bias.

Similarly, the $l$-th decoder layer can be represented by  \cite{ye2019understanding}:
\begin{eqnarray}\label{eq:Dec}
\tilde \z^{l-1}=\sigma\left(\Db^l \tilde\z^{l} \right)
\end{eqnarray}
where 
   \begin{eqnarray}
 \Db^l= \begin{bmatrix} 
\tilde\Phib^l\circledast \tilde\psib^l_{1,1} & \cdots &  \tilde\Phib^l \circledast \tilde\psib^l_{1,q_l}  \\
  \vdots & \ddots & \vdots \\
\tilde\Phib^l\circledast \tilde\psib^l_{q_{l-1},1} & \cdots &  \tilde\Phib^l\circledast \tilde\psib^l_{q_{l-1},q_{l}}
 \end{bmatrix}
 \end{eqnarray}
 where 
$\tilde\Phib^l$ denotes the $m_l\times m_l$  matrix that represents the unpooling operation at the $l$-th layer, and
  $\tilde\psib_{i,j}^l\in \Rd^r$ represents
the $l$-th layer decoder filter to generate the $i$-th channel output from the contribution of the $j$-th channel input.

Next, consider  the skipped branch signal $\chib^l$ by concatenating the output for each skipped branch as shown in Fig.~\ref{fig:linearCNN}(b).
It is easy to show that the $l$-th encoder layer with the skipped connection 
can be represented
by  \cite{ye2019understanding}:
\begin{align}
\begin{bmatrix}\z^l\\ \chib^l \end{bmatrix} 
 &= \begin{bmatrix} \sigma\left(\Eb^{l\top} \z^{l-1}\right)\\  \z^{l-1}\end{bmatrix} \label{eq:skip_enc}\\
 \tilde \z^{l-1} &=\sigma\left(\Db^l \tilde\z^{l} +\Db^l \chib^l\right) \ . \label{eq:skip_dec}
\end{align}
\add{If the skipped branch is concatenated in channel dimension, \eqref{eq:skip_dec} can be 
represented by
\begin{align}
\tilde \z^{l-1} &=\sigma\left(\Db^l  \begin{bmatrix}\tilde\z^{l}  \\ \chib^l\end{bmatrix}\right) 
=\sigma\left(\Db_z^l \tilde\z^{l} +\Db_s^l \chib^l\right) \ , \label{eq:skip_dec1}
\end{align}
where $\Db^l=[\Db_z^l~~\Db_s^l]$ denotes the decoder filter matrix for the channel concatenation, with $\Db_z^l$ and $\Db_s^l$ being the
filter block with respect to $\tilde \z^l$ and the skipped signal $\chib^l$, respectively.
Due to the similarity between \eqref{eq:skip_dec} and \eqref{eq:skip_dec1}, for the rest of the paper we use
\eqref{eq:skip_dec} for simplicity.
}

\subsection{Linear CNN}

First, consider a {\em linear} encoder-decoder CNN {\em without} skipped {connections} as shown in Fig.~\ref{fig:linearCNN}(a), where
there exists  no ReLU nonlinearity. 
In this case,  we have the following linear representation at the $l$-th encoder layer \cite{ye2019understanding}:
\begin{eqnarray}
 \z^l
 = \Eb^{l\top} \z^{l-1}
\end{eqnarray}
Similarly, the $l$-th decoder layer can be represented by
\begin{eqnarray}
\tilde \z^{l-1}=\Db^l \tilde\z^{l} 
\end{eqnarray}
Then, 
 the output $\vb$ of the encoder-decoder CNN  with respect to input $\z$ can be represented by the following  basis-like representation \cite{ye2019understanding}:
\begin{eqnarray}\label{eq:basis}
\vb 
~= \Tc_\Thetab(\zb) = \sum_{i} \left\langle {\blmath b}_i, \z \right\rangle \tilde  {\blmath b}_i
\end{eqnarray}
where  $\Thetab$ refers to all encoder and decoder convolution filters,
and $ {\blmath b}_i$ and $\tilde  {\blmath b}_i$ denote the $i$-th column of the following matrices, respectively:
\begin{eqnarray}\label{eq:B0}
\B = \ \Eb^1\Eb^2 \cdots   \Eb^{\kappa} &,&
\tilde\B= \Db^1\Db^2 \cdots   \Db^{\kappa}
\end{eqnarray}
Note that this representation is completely linear, since the representation does not vary once the network parameters $\Thetab$ are trained.
Furthermore, consider the following {\em frame conditions} for the pooling and filter layers:
\begin{eqnarray}\label{eq:frame}
\tilde\Phib^l \Phib^{l\top}= \alpha \Ib_{m_{l-1}}&,& \Psib^l \tilde\Psib^{l\top} = \frac{1}{r\alpha}\Ib_{rq_{l-1}},\quad \forall l
\end{eqnarray}
where $\Ib_n$ denotes the $n\times n$ identity matrix, $\alpha>0$ is a nonzero constant,
and
\begin{align}
\Psib^l &=\begin{bmatrix} 
 \psib^l_{1,1} & \cdots &  \psib^l_{q_l,1}  \\
  \vdots & \ddots & \vdots \\
 \psib^l_{1,q_{l-1}} & \cdots &\psib^l_{q_{l},q_{l-1}}
 \end{bmatrix},\\
 \tilde\Psib^l&=\begin{bmatrix} 
\tilde\psib^l_{1,1} & \cdots & \tilde\psib^l_{1,q_l}  \\
  \vdots & \ddots & \vdots \\
\tilde\psib^l_{q_{l-1},1} & \cdots & \tilde\psib^l_{q_{l-1},q_{l}}
 \end{bmatrix}
\end{align}
Under these frame conditions,  we showed in  \cite{ye2019understanding} that  \eqref{eq:basis} satisfies the perfect reconstruction condition, i.e
\begin{align}\label{eq:PR}
 \zb = \sum_{i} \left\langle {\blmath b}_i, \z \right\rangle \tilde  {\blmath b}_i,
 \end{align}
hence the corresponding linear CNN is indeed a frame representation, similar to wavelet frames \cite{cai2012image}.

\subsection{Role of Skipped Connection}


Without ReLU,  the decoder branch of the skipped connection in \eqref{eq:skip_dec} can be simplified to a linear sum:

{
\begin{eqnarray}
\tilde \z^{l-1}=\Db^l \tilde\z^{l} +\Db^l \chib^l \ .
\end{eqnarray}
}
This leads to the same expression in \eqref{eq:basis}  except that
 $\bb_i$ and $\tilde\bb_i$ are the $i$-th column of the following augmented
matrices $\B_{skp}$ and $\tilde\B_{skp}$, respectively \cite{ye2019understanding}:
\begin{align}
\B_{skp} = \left[ \Eb^1 \cdots  \Eb^{\kappa},~\underbrace{\Eb^1 \cdots  \Eb^{\kappa-1}\Sb^\kappa, ~\cdots, ~ \Eb^1\Sb^2, ~\Sb^1}_{\text{augmented blocks from skipped connection}} \right]
\end{align}
\begin{align}
\tilde\B_{skp} = \left[ \Db^1 \cdots  \Db^{\kappa},~\underbrace{\Db^1 \cdots  \Db^{\kappa-1}\tilde\Sb^\kappa, ~\cdots, ~ \Db^1\tilde\Sb^2, ~\tilde\Sb^1}_{\text{augmented blocks from skipped connection}} \right]
\end{align}
where
  \begin{eqnarray}
\Sb^l= \begin{bmatrix} 
\Ib_{m_l}\circledast \psib^l_{1,1} & \cdots &  \Ib_{m_l}\circledast \psib^l_{q_l,1}  \\
  \vdots & \ddots & \vdots \\
\Ib_{m_l}\circledast \psib^l_{1,q_{l-1}} & \cdots &\Ib_{m_l}\circledast \psib^l_{q_{l},q_{l-1}}
 \end{bmatrix}
 \end{eqnarray}
  \begin{eqnarray}
\tilde \Sb^l= \begin{bmatrix} 
\Ib_{m_l}\circledast \tilde\psib^l_{1,1} & \cdots &  \Ib_{m_l}\circledast \tilde\psib^l_{1,q_l}  \\
  \vdots & \ddots & \vdots \\
\Ib_{m_l}\circledast \tilde\psib^l_{q_{l-1},1} & \cdots &  \Ib_{m_l}\circledast \tilde\psib^l_{q_{l-1},q_{l}}
 \end{bmatrix}
 \end{eqnarray}
 Furthermore, the frame condition in \eqref{eq:frame} can be slightly modified as
 \begin{align}\label{eq:frame_skip}
\tilde\Phib^l \Phib^{l\top}= \alpha \Ib_{m_{l-1}},&~ \Psib^l \tilde\Psib^{l\top} = \frac{1}{r(\alpha+1)}\Ib_{rq_{l-1}},\quad \forall l
\end{align}
Under this condition, the same perfect reconstruction condition \eqref{eq:PR} holds.
Therefore, the role of the skipped connection is to make the basis representation more redundant.
 Since the redundant basis representation makes the approximation more robust to noise \cite{daubechies1992ten}, we  expect
  that  the skipped connection plays the same role \cite{ye2019understanding}.

\subsection{CNN with ReLU}

In \cite{ye2019understanding} we showed that even with  ReLU nonlinearities  the expression \eqref{eq:basis} is still
valid. The only change is that the basis matrix has additional ReLU pattern blocks in between encoder, decoder, and skipped blocks.
For example, the expression in \eqref{eq:B0} is changed as follows:
\begin{eqnarray}\label{eq:B1}
\B(\z)  &=& \ \Eb^1\Lambdab^1(\z)\Eb^2 \Lambdab^2(\z) \cdots   \Lambdab^{\kappa-1}(\z)\Eb^{\kappa} \\
\tilde\B(\z) &=&  \Db^1\tilde\Lambdab^1(\z)\Db^2 \tilde\Lambdab^2(\z)\cdots \tilde\Lambdab^{\kappa-1}(\z)  \Db^{\kappa}
\end{eqnarray}
where $\Lambdab^l(\z)$ and $\tilde\Lambdab^l(\z)$ are the diagonal matrices
with 0 and 1 elements indicating the ReLU activation patterns. Similar modification
can be added in $\B_{skp}$ and $\tilde \B_{skp}$ \cite{ye2019understanding}.

Accordingly, the linear representation in \eqref{eq:basis}  should be modified as a nonlinear representation:
\begin{eqnarray}\label{eq:basis2}
\vb 
~= \Tc_\Thetab(\zb) = \sum_{i} \left\langle {\blmath b}_i(\z), \z \right\rangle \tilde  {\blmath b}_i(\z)
\end{eqnarray}
where we now have an explicit dependency on $\z$ for ${\blmath b}_i(\z)$ and $\tilde {\blmath b}_i(\z)$ due to the
input dependent ReLU activation patterns, which makes the representation nonlinear.
\add{If each ReLU activation pattern is  independent
to each other}, then the number of distinct ReLU activation pattern is $2^{\text{\# of neurons}}$, where the
number of neurons is determined by the number of the entire features. Therefore, the number of distinct
linear representations increases exponentially with depth,  and width \cite{ye2019understanding}.
\add{In practice, ReLU activation patterns are not completely independent, so that the actual number
of piecewise linear representations can be smaller. This issue will be discussed later.}

\subsection{Deep Learning vs. Basis Pursuit}
\label{sec:CNN vs. Basis Pursuit}
\begin{figure}[!hbt] 	
\center{ 
\includegraphics[width=7cm]{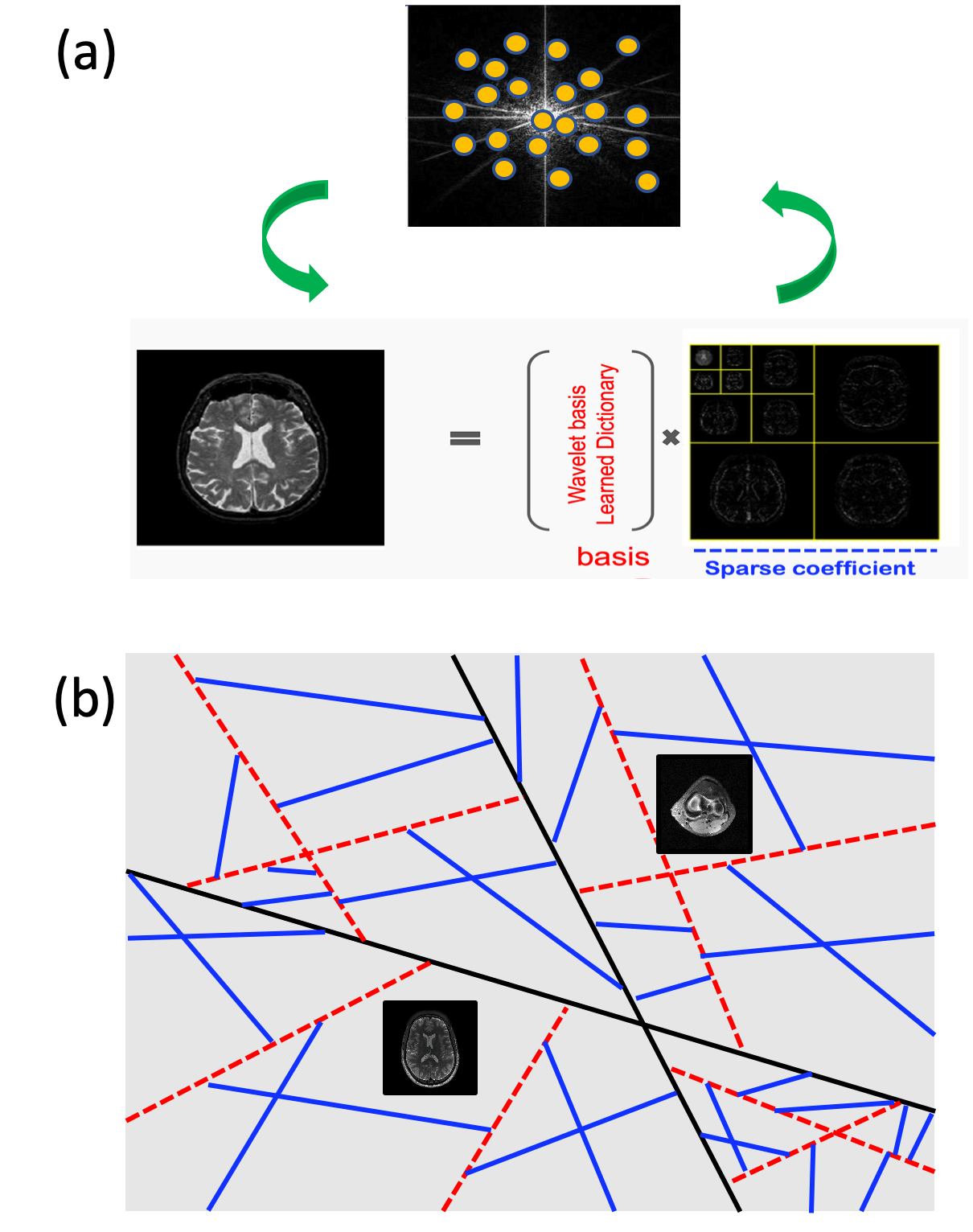}
}
\caption{Reconstruction principle of (a) basis pursuit, and (b) deep neural networks. }
\label{fig:principle}
\end{figure}

\add{
Over the last decade, the most widely used mathematical tool in the signal processing theory were compressed sensing or sparse representation techniques. In particular, the compressed sensing theory is based on the observation that when images are represented via bases of frames, in many cases they can be represented as a sparse combination of bases or frames. Thanks to the sparse representation, even when the measurements are very few below the classical limits such as Nyquist limit, one could obtain a stable solution of the inverse problem by searching for the sparse representation that generates consistent output to the measured data as shown in Fig.~\ref{fig:principle}(a). As a result, the goal of the image reconstruction problem is to find an optimal set of sparse basis function suitable for the given measurement data. This is why the classical method is often called the {\em basis pursuit.} 
}

\add{The deep learning approach and the basis pursuit appear as two completely different approaches. However, \eqref{eq:basis2} showed that indeed there exists a very close relationship between the two. 
Specifically, since nonlinearity is applied after the convolution operation, the on-and-off activation pattern of each ReLU determines a binary partition of the feature space at each layer across the hyperplane that is determined by the convolution. Accordingly, in deep neural networks, the input space is partitioned into multiple non-overlapping regions  so that input images for each region share the same linear representation, but not across the partition. This implies that two different input images are automatically switched to two distinct linear representations that are different from each other as shown in Fig.~\ref{fig:principle}(b).
Hence, although  the deep neural network is indeed similar to the classical basis pursuit algorithm that searches for  distinct linear representations for each input, deep neural networks have an important computational advantage over the classical basis pursuit approaches that rely on computationally expensive optimization techniques.}

\add{ Moreover, the representations are entirely dependent on the filter sets that are learned from the training data, which is different from the classical representation learning approaches that are designed by a top-down model. Furthermore, with more  number of input space partitions and the associated distinct linear representations, the nonlinear function approximation by 
the piecewise linear basis representation becomes more accurate.
Therefore, the number of piecewise linear regions is directly related to the expressivity or representation power of the neural network, and the goal
of this paper is to provide a geometric approach to increase the number of partitions with minimal network complexity increase.
}

\section{Main Contribution}

In this section, which is novel, we provide  further geometric insights that lead to  novel network architecture modification schemes to increase the \add{expressivity of the network} to improve reconstruction performance with
minimal overhead.

\subsection{Geometric Meaning of Features}

One of the most interesting questions about neural network is understanding the meaning of the intermediate features that are obtained as an output
of each layer of neural network.  Although these are largely regarded as {latent variables}, to the best of our knowledge,
the geometric understanding of each latent variable is still not complete.
In this section, we show that this intermediate feature is directly related to the relative coordinates
with respect to the hyperplanes that partition the product space of the previous layer features. 

To understand the claim, let us first revisit the ReLU operation for each neuron at the encoder layer.
Let $\Eb_i^l$ denote the $i$-th column of encoder matrix $\Eb^l$ and $z_i^l$ is the $i$-th element of $\z^l$.
Then,   the output of an {\em activated} neuron can be represented as: 
\begin{align}
z_i^l = \underbrace{\frac{|\langle \Eb_i^l, \zb^{l-1} \rangle|}{\|\Eb_i^l\|}}_{\text{distance to the hyperplane}} \times\quad \|\Eb_i^l\| 
\end{align}
where the normal vector of the hyperplane can be identified as
\begin{equation}
\nb^l=\Eb_i^l.
\end{equation}
This implies that the output of the activated neuron is the scaled version of the distance to the hyperplane
which partitions the space of
feature vector $\zb^{l-1}$ into active and non-active regions.
Therefore, the role of the neural network can be understood as the representing the
input data  using a coordinate vector  using  the relative distances with respect to  multiple hyperplanes.

\begin{figure}[!hbt] 	
\center{ 
\includegraphics[width=8.0cm]{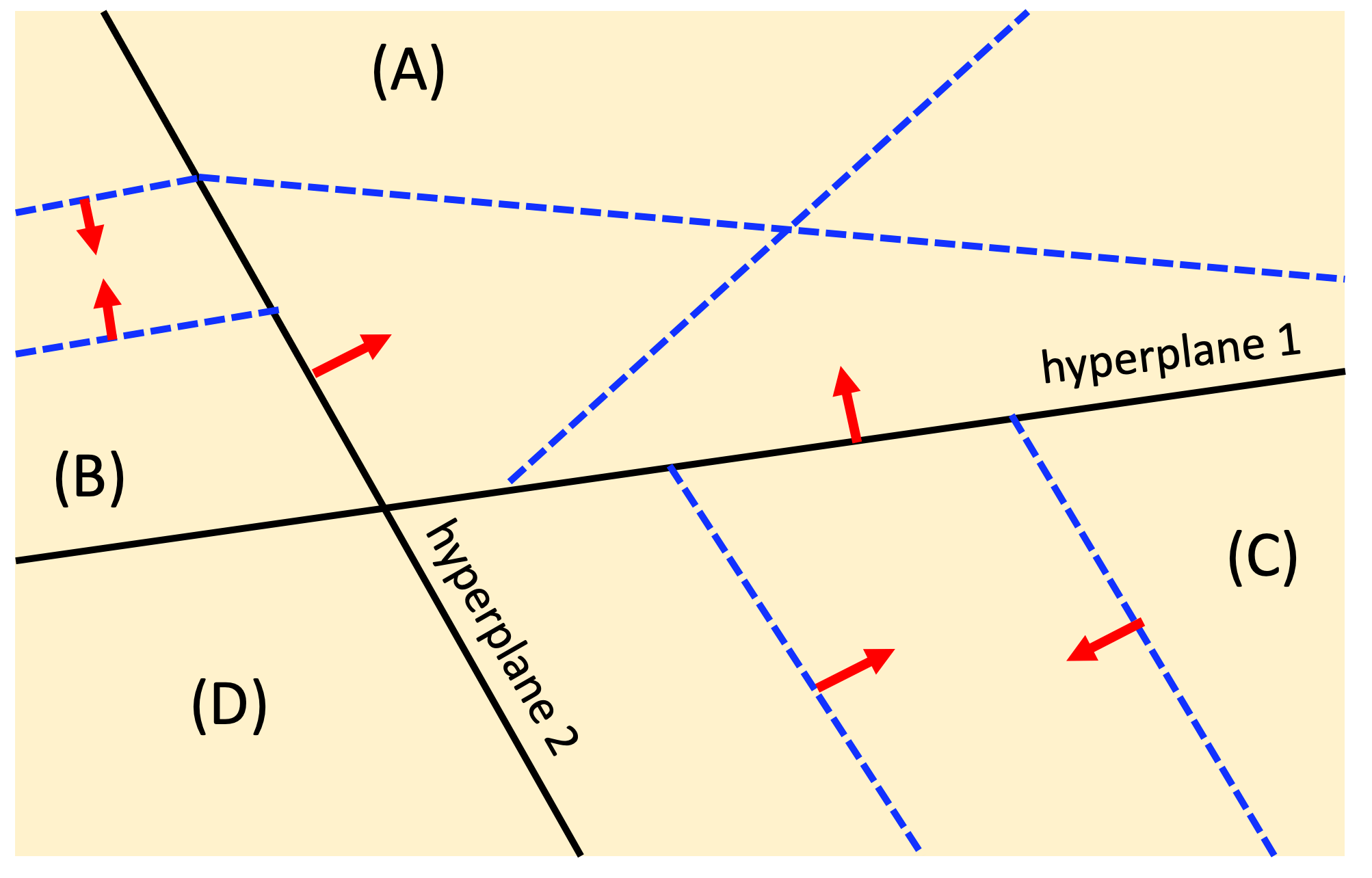}
}
\caption{Two layer neural network with two neurons for each layer.  Red arrows indicate the normal
direction of the hyperplane. The black lines are hyperplanes for the first layers, and the blue lines  correspond to the second layer hyper planes. }
\label{fig:hyperplane}
\end{figure}

In fact, the aforementioned interpretation of the feature may not be novel, since similar interpretation can be used to explain the geometrical
meaning of 
the linear frame coefficients. Instead, one of the most important differences comes from the multilayer representation.
%
%
To understand this, consider the following two layer neural network:
\begin{align}\label{eq:zi}
 z_i^l
 &=  \sigma(\Eb_i^{l\top}\zb^{l-1})
 \end{align}
 where
 \begin{align}
 \zb^{l-1} &= \sigma\left(\Eb^{(l-1)\top} \z^{l-2}  \right) \nonumber \\
 & =  \Lambdab (\z^{l-1}) \Eb^{(l-1)\top} \z^{l-2} 
\end{align}
where $ \Lambdab (\z^{l-1})$ again encodes the ReLU activation pattern. Using the property of inner product and adjoint operator, we have 
\begin{align}
 z_i^l
 &=  \sigma(\Eb_i^{l\top}  \zb^{l-1}) \nonumber \\
 &= \sigma\left( \left\langle \Eb_i^l,  \Lambdab (\z^{l-1}) \Eb^{(l-1)\top} \z^{l-2} \right\rangle \right) \nonumber \\
 &= \sigma\left(  \left\langle  \Lambdab (\z^{l-1}) \Eb_i^l, \Eb^{(l-1)\top} \z^{l-2} \right\rangle \right)
\end{align}
This indicates that on the space of  the {\em unconstrained} feature vector from the previous layer (i.e. no ReLU is assumed),
the hyperplane normal vector is now changed to 
\begin{align}\label{eq:n}
\nb^{l}=\Lambdab (\z^{l-l}) \Eb_i^l.
\end{align}
This implies that the hyperplane in the current layer 
is adaptively changed with respect to the input data, since the ReLU
activation pattern in the previous layer, i.e. $\Lambdab (\z^{l-l})$ can vary
depending on the inputs. This is an important difference over the linear multilayer frame representation, whose hyperplane structure is the same regardless of different inputs.

For example,
Fig.~\ref{fig:hyperplane} shows a partition geometry  of $\Rd^2$ by a two-layer neural network with two neurons at each layer.
The normal vector direction for the second layer hyperplanes are determined
by the ReLU activation patterns such that the coordinate values at the inactive neuron become degenerate. 
More specifically, for the (A) quadrant where two neurons at the first layers are active,  we can obtain two hyperplanes in any normal direction determined
by the filter coefficients.
However,  for the (B) quadrant where  the second neuron is inactive, the situation is different.
Specifically,  due to \eqref{eq:n},  the second coordinate of the normal vector, which corresponds to the inactive neuron, becomes degenerate.
This leads to  two  parallel hyperplanes that are distinct only by the bias term.
Similar phenomenon occurs for the quadrant (C) where 
the first neuron is inactive.
 For the (D) quadrant where two neurons are inactive, the normal vector becomes zero and there exists no partitioning.
Therefore, we can conclude that the hyperplane geometry is adaptively determined by the feature vectors in the previous layer.

\begin{figure}[!hbt] 	
\center{ 
\includegraphics[width=8.0cm]{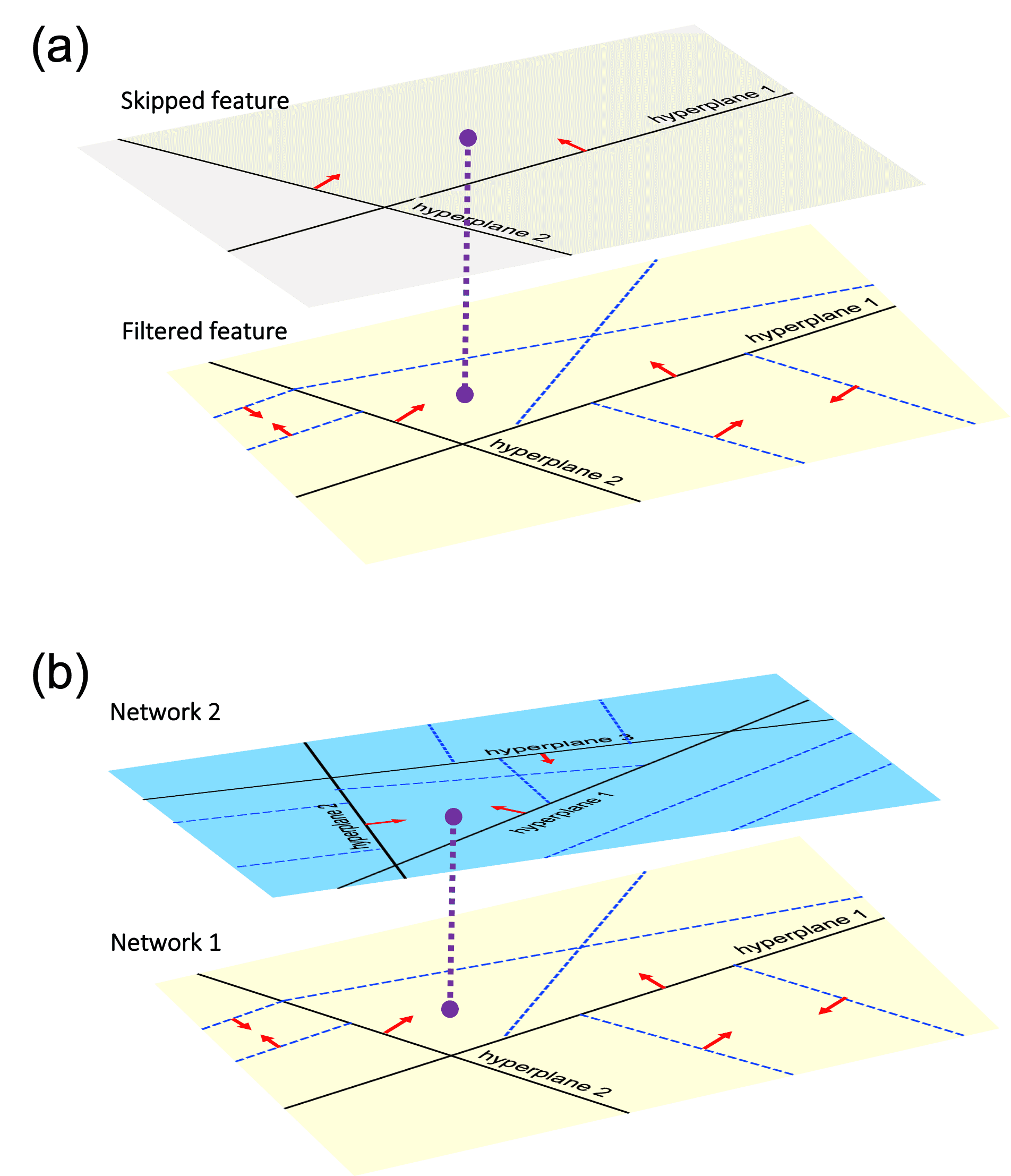}
}
\caption{Partition geometry of (a) skipped connection, and (b) a general expressivity enhancement scheme using two networks.}
\label{fig:skip_partition}
\end{figure}

For the case of the skipped connection,  \eqref{eq:skip_dec} informs that we have
additional features from the previous layer that retains the unprocessed information. This provides redundant
information from distinct partitions as shown in Fig.~\ref{fig:skip_partition}(a).

Finally, for the case of CNN, the choice of the hyperplanes becomes further constrained
due to the convolutional relationship.
For example,  to encoder the data manifold in $\Rd^3$ with the $r=2$ convolution filter with the filter coefficient of $[1,2]$,   
the following three vectors determine the normal direction of the three hyperplanes:
\begin{align}
\nb_1^l=\begin{bmatrix} 1 & 2 & 0 \end{bmatrix}, \quad \nb_2^l=\begin{bmatrix} 0 &1 &2  \end{bmatrix}, \quad\nb_3^l=\begin{bmatrix}2 &  0 & 1 \end{bmatrix}
\end{align}
where we assume the circular convolution and no pooling operation (i.e. $\Phib^l=\Ib_{3}$).
This implies that  each channel of the convolution filter determines an orthant of the underlying feature space, and the
feature vectors are directly related to the coordinate on the resultant orthant.

\subsection{Geometric Meaning of Decoder}

Based on our understanding of multilayer encoder structure, here we also provide
a geometric interpretation of the decoder in a multilayer setup.
Consider the {\em unconstrained} feature before ReLU:
\begin{align}\label{eq:unconst}
\tilde\xib^l := \Db^{l+1} \tilde\z^{l+1}, \quad l\in[\kappa]
\end{align}
 which corresponds to the synthesis
formula from the features $\tilde\z^l$ and the synthesis frame $\Db^l$.
Then, from   \eqref{eq:Dec} we have
\begin{align}\label{eq:DecXi}
\tilde \xib^{l-1}=\Db^l \tilde\zb^l &= \Db^l \sigma\left(\tilde\xib^{l}\right)  \\
&= \left(\Db^l \Lambdab(\tilde\xib^{l})\right)\tilde\xib^{l} 
\end{align}
where $\Lambdab(\tilde\xib^l)$ denotes the ReLU activation patterns from the $l$-th decoder layer.
Accordingly, the operation of the decoder can be interpreted as representing the signal with the unconstrained coefficients $\tilde\xib^{l} $ and the
synthesis frame $\Db^l \Lambdab(\tilde\xib^{l})$.
Since $\Lambdab(\tilde\xib^l)$ is a diagonal matrix with 0 and 1 values depending on the ReLU activation patterns,
the synthesis frame $\Db^l \Lambdab(\tilde\xib^{l})$ is indeed a subset of the original synthesis frame.
Therefore, the ReLU activation patterns work as a subset selection criterion of the original synthesis frame.
This is in contradiction to the classical frame-based signal processing that requires explicit shrinkage operation
to select the optimal frame basis.  In multi-layer neural network, this process can be done automatically
by the ReLU activation patterns.

In addition, the geometry of the decoder with the skipped connection turns out to be more complicated.
From \eqref{eq:skip_dec} and \eqref{eq:unconst}, we have
\begin{align}
 \tilde \xib^{l-1} =\Db^l \tilde\z^{l} +\Db^l \chib^l 
 &= \left(\Db^l \Lambdab(\tilde\xib^{l})\right)\tilde\xib^{l} +\Db^l \chib^l \nonumber \\
 &= \underbrace{\begin{bmatrix} \Db^l \Lambdab(\tilde\xib^{l}) & \Db^l \end{bmatrix}}_{\text{augmented frame basis}} \begin{bmatrix}\tilde\xib^l \\ \chib^l \end{bmatrix}
\end{align}
where the augmented frame basis is composed of the adaptive basis by the ReLU activation
and an {\em unconstrained} basis for the skipped branch. 
Therefore, even though we use the same decoder operation using $\Db^l$, the skipped branch
makes the representation more redundant and the shrinkage behavior in
\eqref{eq:DecXi} is now dependent on the signal in the skipped branch.

\subsection{Expressivity Enhancement}

Based on the discussion so far, we now understand that there is an important mechanism to improve the expressivity of the neural network: one by the increasing the number of piecewise linear regions from the partition, and the other for improving the noise robustness by increasing the redundancy of the basis representation at each piecewise linear regions. The aforementioned discussion has also revealed that the skipped connection achieves both goals with basically no additional complexity thanks to the redundant representation using distinct set of partitions. In this section, we show that
the idea can be generalized to improve the expressivity of the neural network with negligible complexity overhead.

Suppose we are given multiple neural networks $\Tc^{(n)}_{\Theta}$ with enough capacity such that target vector
$\vb$ can be equally represented by
\begin{eqnarray}
\vb =  \Tc^{(n)}_{\Theta} (\z), 
\quad n=1,\cdots, N
\end{eqnarray}
We will defer the discussion as to how such multiple neural networks can be obtained with minimal complexity increase.
We now propose the following nonlinear summation of the result using attention neural network
\begin{align}\label{eq:attention}
\vb  &= \Bc(\wb,\Thetab)(\z):=  \sum_n   w_n(\z) \Tc^{(n)}_{\Theta} (\z)
\end{align}
where $\wb(\z)= \{w_n(\z)\}_{n=1}^N$ refers to the output of the attention neural network.
The specific design of the attention module will be discussed later.

One of the most important advantages of the nonlinear combination in \eqref{eq:attention} is the increased expressivity of the resulting neural network.
This is because distinct piecewise linear region can be
partitioned using  different neural networks as shown in Fig.~\ref{fig:skip_partition}(b). 
 Accordingly,
 the number of piecewise linear regions can be upper-bounded by the product of number of
piecewise linear regions for each neural network.
In addition to the potential increase of the piecewise linear regions, the nonlinear combination model \eqref{eq:attention} increases the redundancy of the
representations in each piecewise linear region. Specifically, using the representation in \eqref{eq:basis} for each neural network,
we have
\begin{align}
\Bc(\wb,\Thetab)(\z) &=  \sum_n   w_n(\z) \Tc^{(n)}_{\Theta} (\z)  \\
&=  \sum_n \sum_{i} w_n(\z)\left\langle {\blmath b}_i^{(n)}(\z), \z \right\rangle \tilde  {\blmath b}_i^{(n)}(\z)  \\
&= \sum_k \left\langle {\blmath b}_k^{at}(\z), \z \right\rangle \tilde  {\blmath b}_k^{at}(\z)
\end{align}
where $\bb_k^{at}$ and $\tilde\bb_k^{at}$ are the $k$-th column of the following concatenated basis matrices, respectively:
\begin{align}
\B^{at}(\z)&= \begin{bmatrix}  \B^{(1)}(\z) & \cdots & \B^{(N)}(\z) \end{bmatrix}  \label{eq:Bat}\\
\tilde\B^{at}(\z)&= \begin{bmatrix} w_1(\z) \tilde\B^{(1)}(\z) & \cdots & w_N(\z)\tilde\B^{(N)}(\z) \end{bmatrix} \label{eq:tBat}
\end{align}
where $\B^{(n)}$ is the concatenation of the $\{\bb_i^{(n)}\}_i$.
Therefore, the representation can become more redundant, and  
the nonlinear combination in \eqref{eq:attention} is expected to improve the reconstruction performance.
In the following, we discuss how to obtain distinct basis representations with minimal complexity increase.

\subsubsection{Bootstrap subsampling}

One simple way to obtain diverse neural network with basically  no complexity increase is the bootstrap and subsampling scheme \cite{breiman1996bagging, cha2019boosting}.
The basic idea is that instead of using all $k$-space measurements, 
 we train the neural network using various subsampling and combine the results using attention.
Note that subsampling  is equivalent to multiplying a  $\{0,1\}$ mask to the input vector.
Therefore, the corresponding neural network output for random sampling pattern can be represented by
\begin{eqnarray}
 \vb 
&=& \sum_{i} \langle {\blmath b}_i(\z), \Mb^{(n)} \z \rangle \tilde  {\blmath b}_i(\x) \\
&=& \sum_{i} \langle \Mb^{(n)} {\blmath b}_i(\z), \z \rangle \tilde  {\blmath b}_i(\x),
\quad n=1,\cdots,N \label{eq:bootstrap}
\end{eqnarray}
where $\Mb^{(n)}$ denotes a  diagonal matrix  with $\{0,1\}$ values depending on the subsampling factor,
and we use the fact that the adjoint of a real diagonal matrix is the original matrix for the second equality.
By inspection, we can easily see that the corresponding sub-basis matrices  in \eqref{eq:Bat}  are given by
\begin{align}
\Bb^{(n)}(\z)= \Mb^{(n)}(\z)  \Bb (\z), 
\quad n=1,\cdots, N. 
\end{align}
As the subsampling patterns are independent to each other, we can easily
see that $\Bb^{(n)}(\z)$ are linearly independent, leading to more redundant representation.
Moreover, although $\tilde\Bb^{(n)}$ are the same as $\tilde\Bb(\z)$ for all $n$, the actual activation patterns and nonzero column of
$\tilde\Bb(\z)$ are determined by $ \langle \Mb^{(n)} {\blmath b}_i(\z), \z \rangle  $, which may lead to distinct representations in the decoder basis.

\begin{figure}[!hbt] 	
\center{ 
\includegraphics[width=8.0cm]{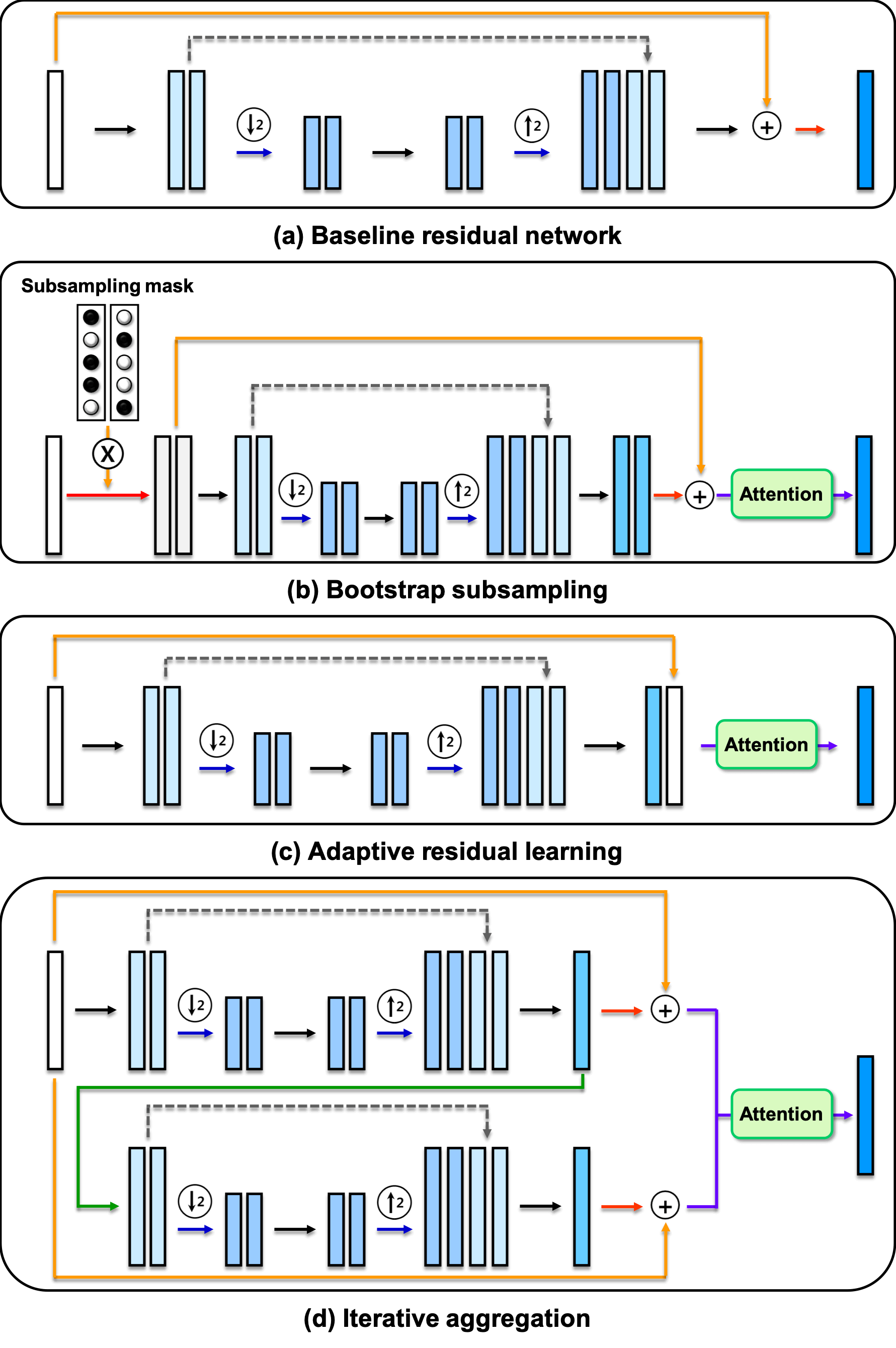}
}
\caption{Simplified flowchart of expressivity enhancement schemes using residual networks. Dashed lines refer to skipped concatenation, and
yellow lines corresponds to the input image copying for residual learning.
(a) Baseline residual network. (b) Bootstrap subsampling with $N=2$, where the network is trained using multiple subsampling and aggregated using an attention network. 
(c) {Adaptive residual learning}, where the input and output of the network were adaptively combined using an attention network. (d) {Iterative aggregation} with $N=2$, where intermediate results are aggregated with an attention
network. } 
\label{fig:boosting}
\end{figure}

\subsubsection{Adaptive Residual Learning}

Residual learning is widely used for medical image restoration \cite{han2016deep, jin2017deep, chen2017low, lee2018deep}, where
the neural network is  trained to learn the difference between the input data $\z$ and the reference data $\vb$. The reconstructed result using residual learning can be represented as follows:
\begin{equation}
\vb = \z + \Tc_\Theta(\z),
\label{eq:residual_output}
\end{equation}
In view of \eqref{eq:attention}, this is equivalent to the sum of the output of identity network and the output of the CNN. 
Therefore, we can generalize the idea to adaptively combine the two network outputs using nonlinear weights:
\begin{equation}
\vb =w_1(\z)\z + w_2(\z)\Tc_\Theta(\z),
\label{eq:adaptive_residual_output}
\end{equation}
which leads to redundant basis representations:
\begin{eqnarray}
\B^{at}(\z)&=& \begin{bmatrix}  \Ib &  \B(\z) \end{bmatrix}  \\
\tilde\B^{at}(\z)&=& \begin{bmatrix} w_1(\z) \Ib &  w_2(\z)\tilde\B(\z) \end{bmatrix}
\end{eqnarray}
Again both the encoder and the decoder basis are distinct, which leads to redundant representations.

\subsubsection{Iterative Aggregation}
Rather than having a parallel branch, here we show that applying the same neural network
multiple times and combining each intermediate results can significantly improve the reconstruction performance with minimal network complexity overhead. Specifically, the main idea is given by
\begin{align}\label{eq:attentioni}
\vb  &= \Bc(\wb,\Thetab)(\z):=  \sum_n   w_n(\z) \overbrace{\left(\Tc_{\Theta} \circ \cdots \circ \Tc_{\Theta}\right)}^{\text{$n$-times}} (\z)
\end{align}
Then, the corresponding sub-basis representation is given by
\begin{align}
 \B^{(n)}(\z) &= \B(\z) \overbrace{\cdots  \tilde\B(\z) \B^{\top}(\z)}^{\text{multiplied by $n-1$ times}}\\ 
\tilde \B^{(n)}(\z)&= w_n(\z)\tilde\B(\z) 
\end{align}
Similar to the bootstrap subsampling, $\tilde\Bb^{(n)}$ are the same as $\tilde\Bb(\z)$ for all $n$, but
the actual activation patterns and nonzero columns of
$\tilde\Bb(\z)$ are determined by $ \langle \ {\blmath b}_i^{(n)}(\z), \z \rangle  $, which may lead to distinct representations in the decoder basis.

Note that in all three methods, the additional complexity comes only from the attention module, which is negligible compared to other network components.
The idea of these expressivity enhancement schemes are illustrated in Fig.~\ref{fig:boosting}.

\section{Method}
\label{method}

\subsection{Training dataset}
We used two types of MR data sets $-$ structural brain images from the Human Connectome Project (HCP) MRI dataset (http://db.humanconnectome.org) and knee images from fastMRI dataset \cite{zbontar2018fastmri}.

The HCP dataset was acquired by Siemens 3T MR system using a 3D spin-echo sequence. The acquisition parameters were as follows: the size of acquisition matrix = 320$\times$320, repetition time (TR) = 3200 ms, echo time (TE) = 565 ms, and field of view (FOV) = 224$\times$224 mm. Among the total of 119 subject data sets, 111 subject data sets  were used for training and validation. The remaining data sets were used for testing. To generate multi-channel $k$-space data, coil sensitivity maps were estimated from a 32$\times$32 block data at the center of $k$-space using MRI simulator (http://bigwww.epfl.ch/algorithms/mri-reconstruction). The number of coils was 16. The fully acquired $k$-space were retrospectively under-sampled by
factors of   3$\times$3 and 4$\times$4 along $k_y \times k_z$ direction, respectively. We included 36$\times$36 low frequency regions for both subsampling cases, which could be used
as the auto calibrating signal (ACS) region for Generalized autocalibrating partially parallel acquisitions (GRAPPA) \cite{griswold2002generalized}.  Accordingly, the net acceleration factors were about $\times$8.12 and $\times$15.42, respectively. 
The readout direction is $k_x$, which is  fully sampled.

The fastMRI dataset was obtained with clinical 3T systems (Siemens Magnetom Skyra, Prisma and Biograph mMR) or 1.5T Siemens Magnetom Area using 2D TSE protocol. The acquisition parameters were as follows: the size of acquisition matrix = 320$\times$320, TR = 2200$\sim$3000 ms, TE = 27$\sim$34 ms, and slice thickness = 3 mm. We used multi-channel dataset with the number of coils of 15. Among the 413 cases of knee data, 402 cases were used for training, 1 case for validation, and the rest for testing. 
 For fastMRI dataset, the downsampling ratio is 4 along the $k_y$ direction with 30 lines of ACS region. The corresponding acceleration ratio was approximately 3.25.

\subsection{Network Architecture}

\begin{figure}[!hbt] 	
\center{ 
\includegraphics[width=8.7cm]{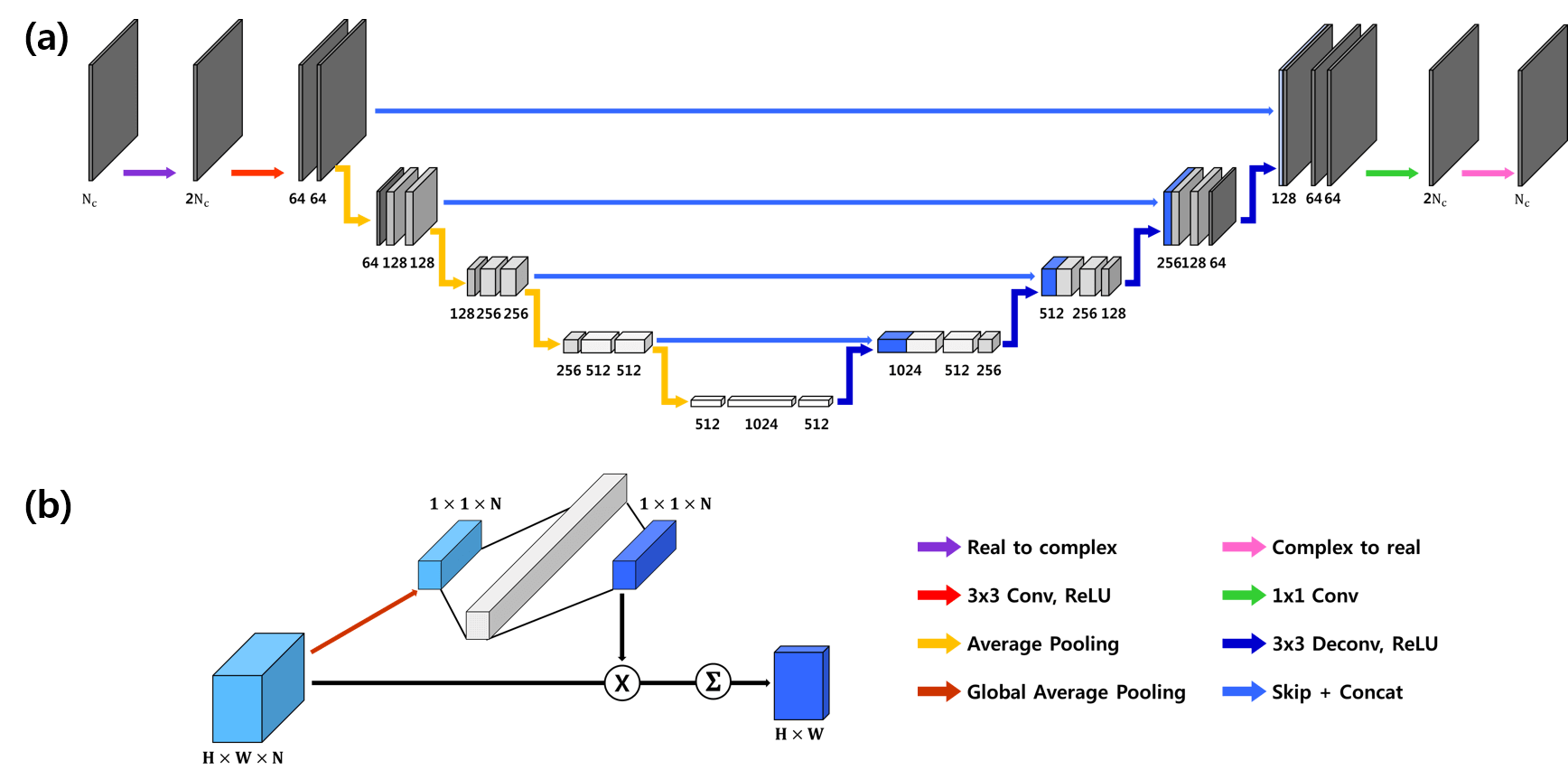}
}
\caption{(a) U-net architecture as the baseline network. (b) Attention network for bootstrap subsampling using an MLP. }
\label{fig:network}
\end{figure}

We employed the U-Net \cite{ronneberger2015u} as the baseline network. 
In particular, we use the residual learning scheme as in Fig.~\ref{fig:boosting}(a),
where the original U-Net part estimate the difference between the input and target data
\cite{ lee2017deep}. 
Specifically, U-net consists of convolution, batch normalization, ReLU, pooling layer, and bypass connection as shown in Fig. \ref{fig:network}(a). To deal with the complex data, there were extra layers for the input and output of the network, which was converted to complex and real value. More specifically, the complex data can be transformed to the real-valued data by concatenation of the real part and the imaginary part of the data along the channel dimension. Therefore, the number of input channels for U-net is 2$\times N_c$, where $N_c$ denotes the number of coils. The complex result can be formed using the real and imaginary channels of the output. 
Each stage contains three sequential layers consisting of 3$\times$3 convolution layer, batch normalization, and ReLU layer, which is presented as a red arrow in Fig. \ref{fig:network}(a). The final layer is 1$\times$1 convolution layer, which is presented by a green arrow. The yellow arrow is 2$\times$2 average pooling. 2$\times$2 unpooling layer was replaced by 3$\times$3 deconvolution layer followed by batch normalization and ReLU. The number of channels increases from 64 in the first stage to 1024 in the final stage. 
\add{Here, the deconvolution layer is implemented as the decoder filtering followed by unpooling layer.}

We evaluated two types of attention networks. 
The attention network for {bootstrap subsampling} scheme  is a multilayer perceptron (MLP) in Fig. \ref{fig:network}(b)\add{, which is a variant of squeeze-and-excitation block  \cite{hu2018squeeze}}. The input of the attention network is $N$ outputs of the  neural network for each subsampling patterns, which
are shrunk to a $N$-dimensional vector through global average pooling. The number of hidden units of MLP is 64. The sigmoid function was applied to the output of MLP, which results in the final  weight vector, $\wb(\zb)\in \Rd^N$. The final output was provided by weighted summation as shown in \eqref{eq:attention}.
For {adaptive residual learning} and {iterative aggregation}, the attention network consists of 1$\times$1 convolution layer that is applied to concatenated intermediate results:
\begin{align}
\vb_{res}^{inter} & = [\zb \quad \Tc_\Theta(\zb) ], \label{eq:res_intermediat_result}\\
\vb_{iter}^{inter}  & = [\Tc_\Theta(\zb) \quad \cdots \quad \overbrace{\left(\Tc_{\Theta} \circ \cdots \circ \Tc_{\Theta}\right)}^{\text{$N$-times}} (\zb) ] ,
\label{eq:iter_intermediat_result}
\end{align}
where $N=4$ is used for our experiment.

\subsection{Network Training}
For  training of the {bootstrap subsampling} scheme, the subsampling process is required. The number of bootstrap subsampling  $N$ for HCP dataset was 10. The same ACS region was used, and the subsampling was randomly performed at 91$\%$ of the original undersampling ratio. The label image was obtained from the fully acquired $k$-space data.  The number of subsampling mask for fastMRI dataset was set to 4. 92$\%$ of the original undersampled position was randomly selected for each bootstrap subsampling with equivalent ACS region. 
%
\add{For training of the {iterative aggregation} scheme, the $n$-th intermediate result $\overbrace{\left(\Tc_{\Theta} \circ \cdots \circ \Tc_{\Theta}\right)}^{\text{$n$ times}} (\zb) $ can be obtained by applying the same neural network, $\Tc_{\Theta}$, $n-$times.}

\add{Our networks were trained to minimize the loss as follows:
\begin{equation}
\min_\Thetab \frac{1}{T}\sum_{t=1}^T ||\vb^{*}_i - \vb_i||_2 = \frac{1}{T}\sum_{t=1}^T ||\vb^{*}_i - \Bc(\wb,\Thetab)(\z_i)||_2,
\end{equation}
where $\vb^*$ is the label image and $\z$ is the input data for network. Depending on training scheme to be used, the output of the neural network can be obtained by \eqref{eq:bootstrap}, \eqref{eq:adaptive_residual_output}, and \eqref{eq:attentioni}.}
We used Adam optimization \cite{kingma2014adam} with the momentum $\beta_1 = 0.9$ and $\beta_2 = 0.999$ for network training. The initial learning rate was set to $10^{-2}$ and $10^{-3}$ for HCP dataset and fastMRI dataset, respectively. The learning rate was halved until it reached around $10^{-4} $ at every 50 and 25 epochs for HCP dataset and fastMRI dataset, respectively. The size of mini-batch was 1.  
For fastMRI dataset, each slice data was divided by the standard deviation of  its absolute value for stable training.
The networks were implemented in Python using TensorFlow library \cite{abadi2016tensorflow} and trained using NVidia GeForce GTX 1080-Ti graphics processing unit and i7-4770 CPU.
The number of training epochs for the HCP dataset was 150, and it took about 10 days. For the fastMRI dataset, the network training took about 23 days, and the number of epochs was 125.  We trained the proposed boosting schemes both in $k$-space domain and image domain.

\add{At the inference stage, the square root of sum of squares (SSoS) images were generated from the reconstructed images of multi-channel data.}

\begin{figure*}[t] 	
\center{ 
\includegraphics[width=19cm]{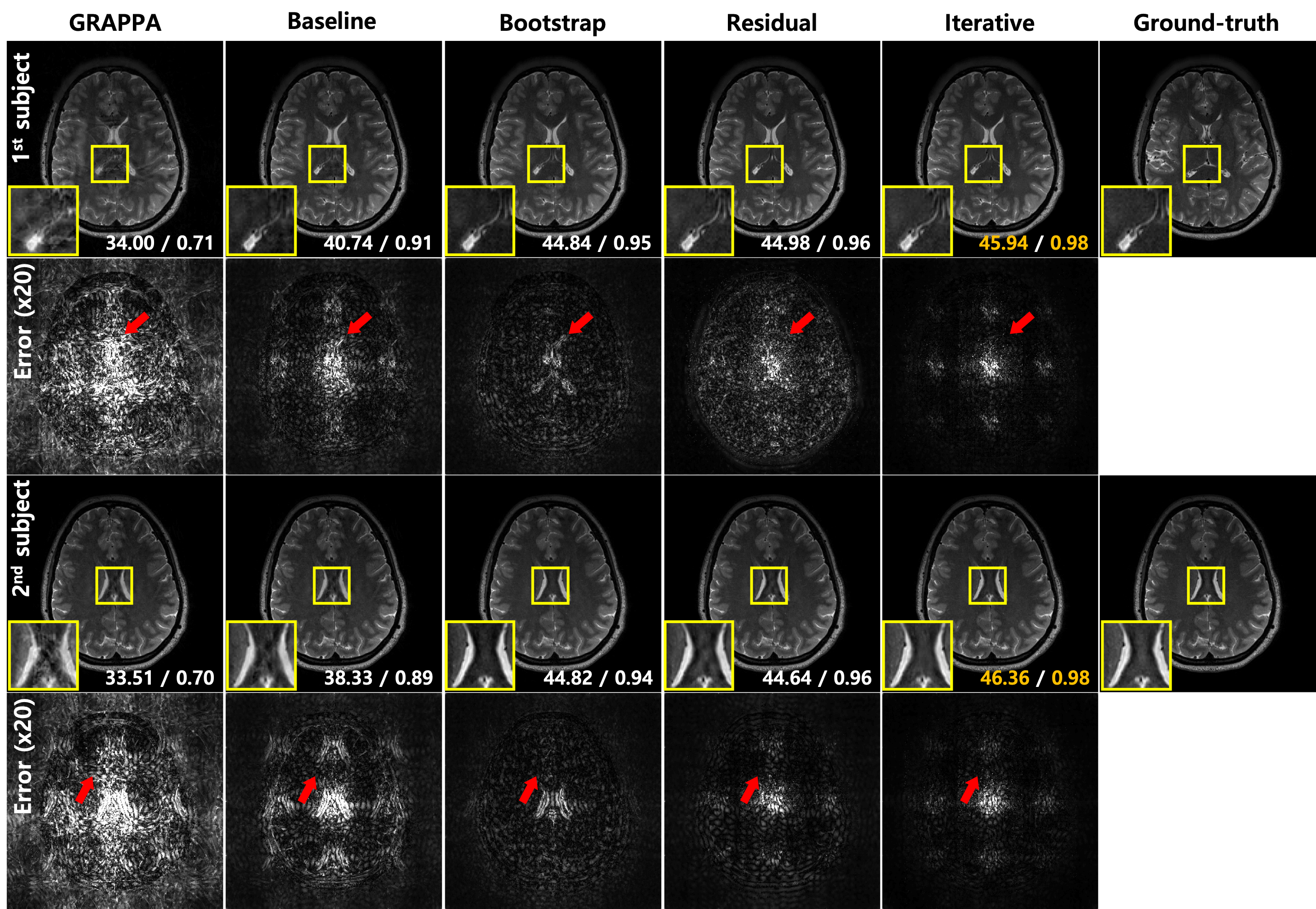}
}
\caption{Reconstruction results using GRAPPA, baseline $k$-space U-Net, {bootstrap subsampling}, {adaptive residual learning}, and {iterative aggregation} at $R=15.42$ of 2-D uniform sampling pattern.  The second and forth rows show the difference images between the reconstructed images and ground-truth images. The numbers written in the images are the corresponding PSNR / SSIM index. }
\label{fig:HCP_recon}
\end{figure*}

\subsection{Comparative studies}
GRAPPA \cite{griswold2002generalized} and  Scan-specific robust artificial-neural-networks for k-space interpolation  (RAKI) \cite{akccakaya2019scan} were used as representative scan-specific reconstruction methods for comparison. \add{We also used the baseline U-Net for image domain learning
and $k$-space learning. The $k$-space learning is a recent state-of-the art deep learning method implemented in $k$-space domain
\cite{han2019k,lee2019k}.}
The parameters for GRAPPA were chosen to provide the best results. The GRAPPA kernel size was 4$\times$4 and 4$\times$1 for the HCP dataset and fastMRI dataset, respectively.

The network used for RAKI is composed of three layers. Each layer has a convolutional layer and nonlinear ReLU operation except for the last layer. 
The baseline RAKI was designed for $k$-space with 1-D  sampling pattern. Therefore, in this experiment
 RAKI was 
only applied for fastMRI data set that are based on 1-D downsampling patterns.
For RAKI training, the $k$-space data was normalized such that the maximum magnitude was set to $0.015$. 
The network was trained using Adam optimization \cite{kingma2014adam}. The learning rate was set to provide the best results.
Each model was trained with 1000 epochs, and the network was retrained for every undersampled $k$-space data.

%


For quantitative evaluation, the peak signal-to-noise ratio (PSNR) and structural similarity (SSIM) \cite{wang2004image} index were used. The PSNR is defined using mean squared error ($MSE$) as:
\begin{equation}
PSNR = 20 \cdot log_{10} \bigg( \frac{MAX_{\xb}}{\sqrt{MSE({\xb}, \xb_*)}} \bigg),
\label{eq:psnr}
\end{equation}
where $\xb$ and $\xb_*$ is the reconstructed SSoS image and ground-truth SSoS image, respectively. $MAX_{\xb}$ denotes the maximum value of the ground-truth SSoS image. SSIM index is one of the perceptual metric, which is defined as
\begin{equation}
SSIM = \frac{(2\mu_{\xb} \mu_{\xb_0} + c_1) (2\sigma_{\xb \xb_* }+ c_2)}{(\mu_{\xb}^2 + \mu_{\xb_*}^2 + c_1)(\sigma_{\xb}^2 + \sigma_{\xb_*}^2 + c_2)},
\label{eq:ssim}
\end{equation}
where $\mu_S$  and $\sigma_S$ denote the mean and the standard deviation of $S$, $\sigma_{ST}^2$ denotes the covariance of $S$ and $T$,  $c_1$ and $c_2$ are the variables to stabilize the division such as $c_1=(k_1L)^2$ and $c_2=(k_2L)$. $L$ is the dynamic range of the pixel intensities. The value of $k_1$ and $k_2$ are followed as the default such as $k_1=0.01$ and $k_2=0.03$.

 \begin{figure*}[!hbt] 	
\center{ 
\includegraphics[width=18cm]{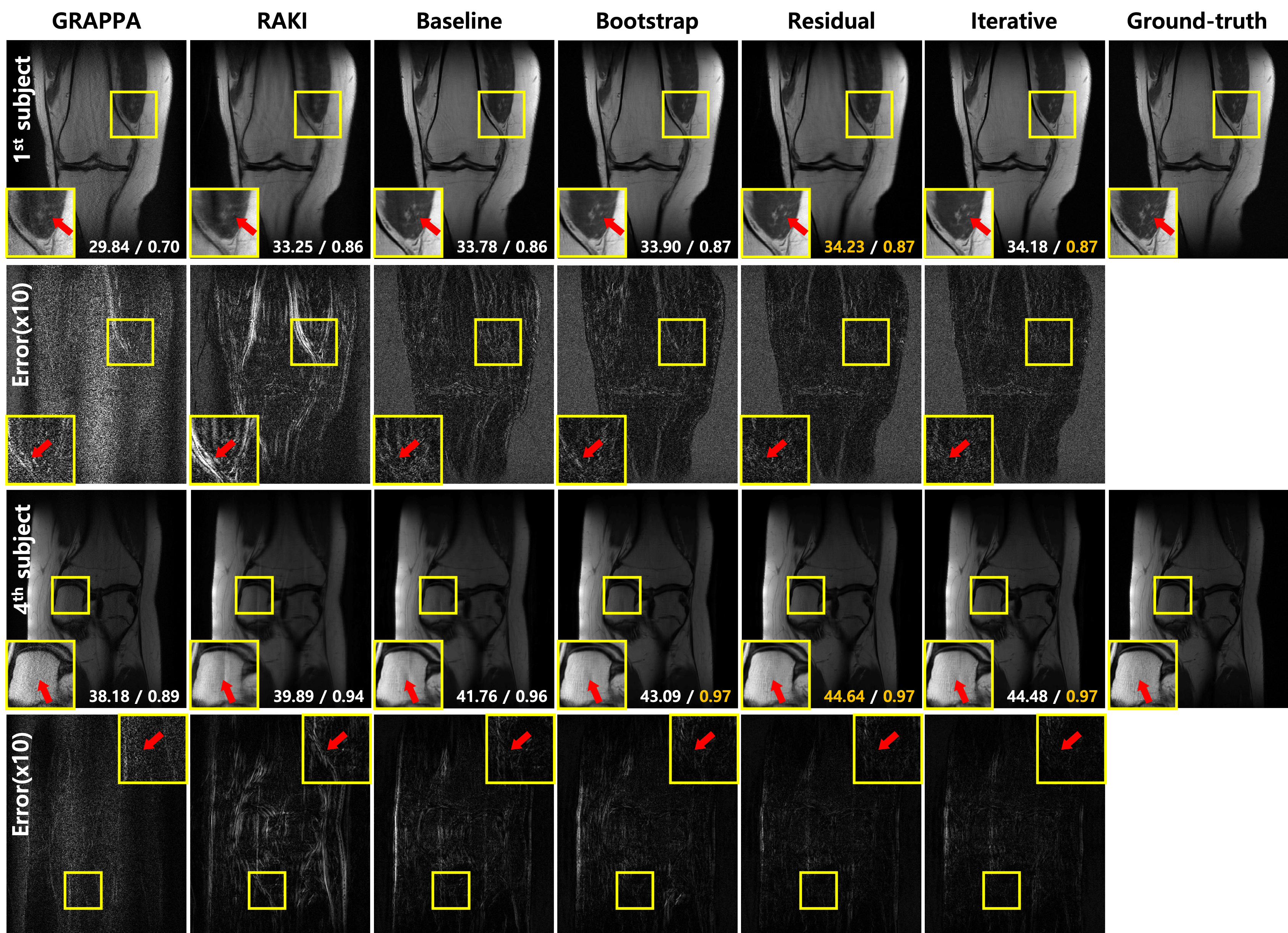}
}
\caption{Reconstruction results using GRAPPA, RAKI, image domain U-Net, {bootstrap subsampling}, {adaptive residual learning}, and {iterative aggregation} at \add{$R=3.25$}. The expressivity enhancement schemes are implemented in the image domain. The yellow boxes demonstrate enlarged images. The numbers written in the images are the corresponding PSNR / SSIM index.}
\label{fig:fastMRI_recon}
\end{figure*}

\section{Experimental results}
\label{results}

\subsection{Simulation results}
First, we trained the proposed model using HCP MR data. In this simulated experiment, eight subjects from HCP dataset were used to verify the performance of the proposed expressivity enhancement schemes for 2-D uniform sampling pattern.
We conducted the experiments at an acceleration factor $R=15.42$.
 The performance of the proposed method was compared with that of GRAPPA,  and baseline U-Net  with   $k$-space learning \cite{han2019k,lee2019k}. 
 
 Fig. \ref{fig:HCP_recon} shows that the aliasing artifacts are present in the reconstructed images using GRAPPA due to the
 calculation of  the kernel using $k$-space information only at the ACS region. 
 Although the baseline $k$-space learning  scheme using U-Net
 provided better reconstruction results by removing many aliasing artifacts, there were still remaining artifacts especially at the center of the image.  
 In contrast, the proposed modification of the $k$-space U-Net significantly reduced the artifacts.  Specifically, the PSNR value of the reconstructed image using the iterative aggregation scheme for the second subject  increased up to 8.03 dB compared to the baseline $k$-space learning as shown in Fig. \ref{fig:HCP_recon}.
 Moreover, the qualitative evaluation also confirmed the quantitative performance improvements.
 

\begin{table}[!hbt]
 \begin{center}
  \begin{small}
 \begin{tabular}{c|c|c|c}
  \toprule
\multicolumn{2}{c|}{}  & PSNR [dB] & SSIM index \\
\midrule
\midrule
\multicolumn{2}{c|}{GRAPPA \cite{griswold2002generalized}}& 34.521 & 0.721 \\
 \midrule
\multirow{4}{*}{Image domain} & Standard & 39.949 & 0.947 \\
& Bootstrap & 40.614 & 0.957 \\
& Residual  & 40.455 & 0.965 \\
& Iterative & 40.330 & 0.954 \\
\midrule
\multirow{4}{*}{$k$-space domain} & Standard & 39.675 & 0.895 \\
&  Bootstrap & 42.719 & 0.924 \\
& Residual  &42.604  & 0.948 \\
& Iterative & \textbf{43.826} & \textbf{0.968} \\
\bottomrule
 \end{tabular}
 \end{small}
 \end{center}
 \caption{Quantitative comparison of various reconstruction methods using HCP dataset at $R=15.42$. The results are mean values from 8 patient data.}
 \label{table:compare_HCP}
 \end{table}

Table \ref{table:compare_HCP} shows average results for the quantitative comparison using 8 patient data set at the test phase.
The proposed expressivity enhancement schemes implemented in both image and $k$-space domain
significantly outperformed GRAPPA, and the $k$-space and image-domain learning \add{using  the baseline U-Net} in terms of PSNR and SSIM index. 
 More specifically, the proposed methods resulted in about 5.7$\sim$9.3 dB gain over GRAPPA. 
 This is because the estimation of GRAPPA kernel becomes more difficult at higher acceleration factors,
 and these limitations can  be reduced using $k$-space deep learning approaches,
since the networks were trained to consider the whole $k$-space region to estimate the interpolation kernel. 
 In addition, our method outperformed the baseline network in both image and $k$-space domain by about 0.6$\sim$4.2 dB in terms of PSNR. This performance gain clearly confirmed the importance of expressivity enhancement  thanks to the bootstrapping and subnetwork aggregation. 

\subsection{In Vivo results}

Fig. \ref{fig:fastMRI_recon} shows reconstruction results at acceleration factor $R=3.25$. Here,
the baseline neural network was the image domain U-Net \cite{lee2017deep,jin2017deep, han2017deep,han2016deep}, and its expressivity enhancement scheme was also implemented in the image domain.
 As shown in Fig. \ref{fig:fastMRI_recon}, the results using GRAPPA were noisy compared to other algorithms. Since the GRAPPA kernel was calculated from the ACS region, the GRAPPA reconstruction is sensitive to measurement noises when the ACS region is not sufficiently large.  Thanks to the nonlinear estimation process, the results using RAKI were better than those using GRAPPA. However, the reconstructed images using RAKI were still blurry compared to the baseline image-domain U-Net and their expressivity enhancement schemes.
\add{In contrast to the baseline U-Net and the proposed schemes that were trained using various training data set,
the representation of RAKI is only determined from the scan-specific ACS data, which leads to relatively limited representation power and poor performance.} 
For example, the detailed structure was not distinguishable using RAKI reconstruction (see red arrow in the result of the first patient in Fig. \ref{fig:fastMRI_recon}), whereas
the proposed expressivity enhancement schemes provided more realistic reconstructed imaged and the detailed structures. 
Among the various expressivity enhancement scheme,
{adaptive residual learning} significantly outperformed the baseline network in terms of PSNR and SSIM values.  Specifically, the PSNR of the reconstructed image using  the {adaptive residual learning} for the fourth subject in Fig. \ref{fig:fastMRI_recon} increased about 2.88 dB.

\begin{table}[h]
 \begin{center}
  \begin{small}
\begin{tabular}{c|c|c|c}
  \toprule
\multicolumn{2}{c|}{}  & PSNR [dB] & SSIM index \\
\midrule
\midrule
\multicolumn{2}{c|}{GRAPPA \cite{griswold2002generalized}}& 30.816 & 0.726 \\
\multicolumn{2}{c|} {RAKI \cite{akccakaya2019scan}} & 32.662 & 0.826 \\
 \midrule
\multirow{4}{*}{Image domain} & Standard & 32.798  & 0.828 \\
&  Bootstrap & 32.973 & 0.830 \\
& Residual  & 33.475 & 0.831 \\
& Iterative & \textbf{33.487} & \textbf{0.832} \\
\midrule
\multirow{4}{*}{$k$-space domain} & Standard & 32.852 & 0.828 \\
& Bootstrap & 33.050 & 0.828 \\
& Residual  & 33.240 & \textbf{0.832}  \\
& Iterative & 33.220 & 0.831 \\
\bottomrule
 \end{tabular}
 \end{small}
 \end{center}
 \caption{Quantitative comparison of various reconstruction methods using fastMRI dataset at $R=3.25$. The values are means values from 10 patient data.}
 \label{table:compare_fastMRI}
 \end{table}
 
Table~\ref{table:compare_fastMRI} shows the average results for quantitative comparison using  10 patient fastMRI data set at the test phase. The quantitative results in Table \ref{table:compare_fastMRI}  confirmed our finding in visual quality evaluation.
Specifically, the proposed method outperformed the baseline image domain and $k$-space learning by about 0.18$\sim$0.69 dB in terms of PSNR.

\subsection{Computation Time}

The computation time per slice was 2.43, 5.24, 0.37, 0.45, 0.39, and 0.47 seconds for GRAPPA, RAKI, baseline neural network, {bootstrap subsampling}, 
{adaptive residual learning}, and {iterative aggregation}, respectively. RAKI took longer
time for the computation, since  the network should be retrained for each patient data. The computation time increase from the proposed expressivity enhancement schemes are negligible and all the expressivity enhancement scheme. Baseline neural networks
were about  5.4 $\sim$ 6.2 times faster than GRAPPA. 
 
\section{Discussion}
\label{discussion}

\subsection{Role of attention module}

 \begin{figure}[h] 	
\center{ 
\includegraphics[width=8cm]{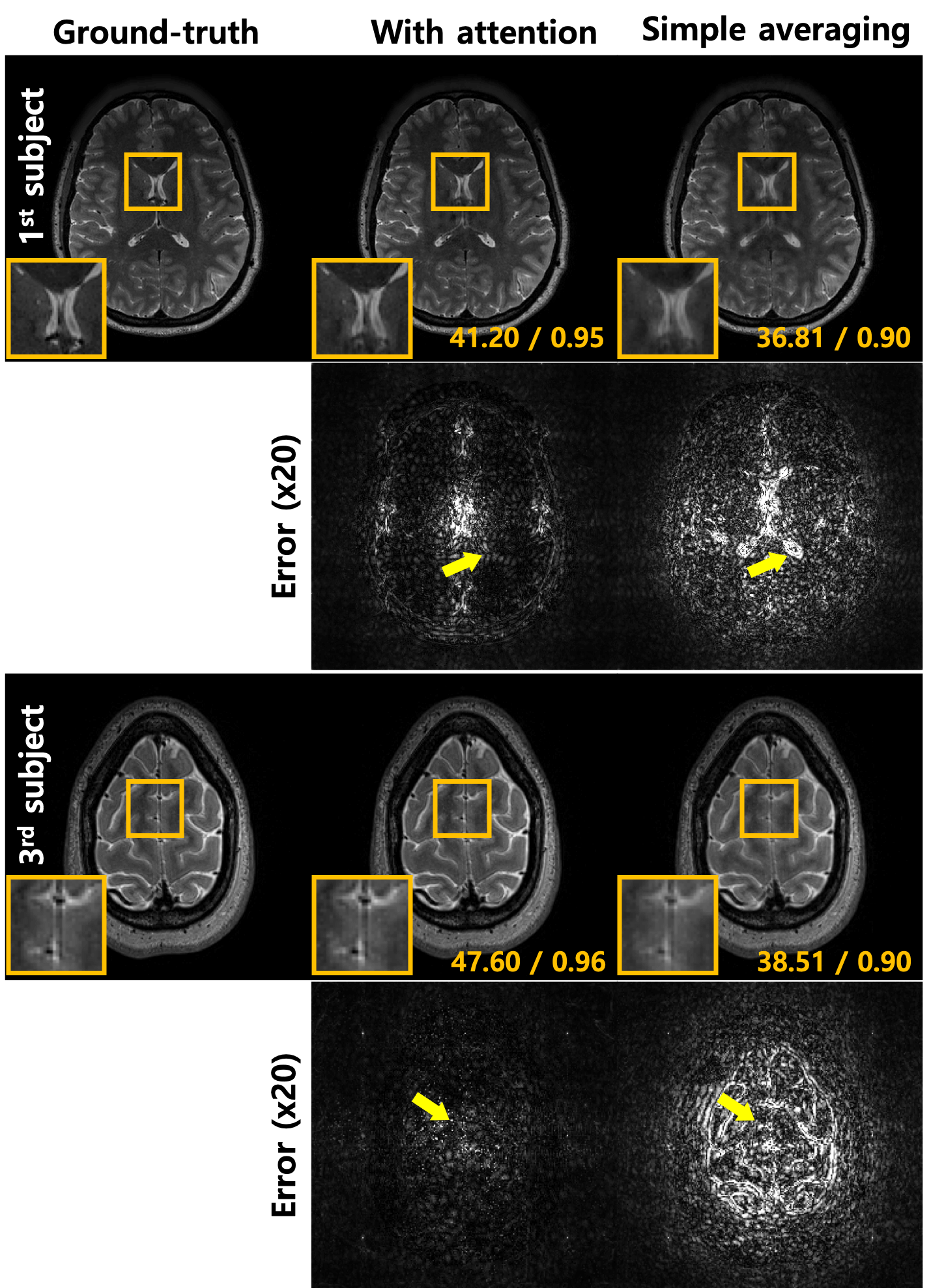}
}
\caption{Effects of attention module for {bootstrap subsampling} scheme at $R=15.42$.}
\label{fig:HCP_attention}
\end{figure}

In order to confirm the role of the nonlinear aggregation using the attention module, we compared the performance with and without attention module for the case of {bootstrap subsampling}. For this purpose, the HCP dataset with $R=15.42$ was used. Without the adaptive weights, $\wb(\zb)$, the final reconstructed result was simply obtained by the average of the overall intermediate results in the absence of the attention module. The network without the attention module resulted in performance drop. Even though this network provided better results than the baseline network, the resulting performance values 
are lower compared to the original performance metric. As shown in Fig. \ref{fig:HCP_attention}, the attention module can reduce the error by efficiently estimating  the weights $\wb(\zb)$. 
This result clearly shows the importance of the attention module. 

\add{
\subsection{Ablation study}
 Our geometric analysis showed that redundant representation leads to an increase in the expressivity of the network. To empirically verify our theoretical findings, we conducted ablation studies using the HCP dataset with $R = 15.42$. First, we performed a  comparative study with different numbers of subsampling patterns ($N$) - 1, 4, 6, and 10 for bootstrap subsampling scheme. As shown in Table \ref{table:ablation}, the quantitative measures were subsequently enhanced with increasing $N$, which shows a direct connection between the network expressivity and the performance. A similar tendency was found in another ablation study with different iterations ($N$) for iterative aggregation scheme. The PSNR value at $N=10$ was 43.826 dB, which is an improvement of about 2.5 dB compared to the baseline network. 
  This clearly confirms the correlation between the redundancy in the basis representation and the expressivity of the network.
 } 

\begin{table}[h]
 \begin{center}
  \begin{small}
\begin{tabular}{c|c|c|c}
  \toprule
& $N$ & PSNR [dB] & SSIM index \\
\midrule
\midrule
Baseline & 1 & 39.675 & 0.895 \\
 \midrule
\multirow{3}{*}{Bootstrap} & 4 & 40.599 & 0.914 \\
&  6 & 41.233 & 0.920 \\
& 10  & \textbf{42.719} &\textbf{0.924} \\
\midrule
\multirow{3}{*}{Iterative} & 2 & 41.348 & 0.946  \\
& 3  & 42.153 & 0.948  \\
& 4 & \textbf{43.826} & \textbf{0.968} \\
\bottomrule
 \end{tabular}
 \end{small}
 \end{center}
 \caption{\add{Quantitative comparison using PSNR and SSIM index with respect to the number of subsampling patterns and iterations for bootstrap subsampling and iterative aggregation scheme, respectively. The results are means values from 8 patient data of the HCP dataset with $R=15.42$.}}
 \label{table:ablation}
 \end{table}

\subsection{Trade-off between expressivity and complexity}

Recall that our expressivity enhancement schemes require more layers and attention module to aggregate the intermediate results. This leads to
additional overhead in terms of trainable parameters, so we are interested in investigating the trade-off between the expressivity and complexity of the network. 

\begin{table}[h]
 \begin{center}
  	\resizebox{0.5\textwidth}{!}{
\begin{tabular}{c|c|c|c|r}
  \toprule
\multicolumn{2}{c|}{}  & Attention  & No. of param. & Overhead (\%)   \\
\midrule
\midrule
\multirow{4}{*}{HCP} & Baseline & $-$ & 25,154,336 &    \\
& Bootstrap & MLP & 25,155,690  & $5.4\times 10^{-3}$ \\
& Residual  & 1$\times$1 Conv & 25,156,416 & $8.3\times 10^{-3}$\\
& Iterative & 1$\times$1 Conv & 25,158,464   &$16.4\times 10^{-3}$ \\
\midrule
\multirow{4}{*}{FastMRI} & Baseline & $-$ & 25,153,054  &  \\
& Bootstrap & MLP & 25,153,634 & $2.3\times 10^{-3}$  \\
& Residual  & 1$\times$1 Conv& 25,154,884 & $7.3\times 10^{-3}$ \\
& Iterative & 1$\times$1 Conv & 25,156,684  & $14.4\times 10^{-3}$  \\
\bottomrule
 \end{tabular}
 }
 \end{center}
 \caption{Network complexity versus PSNR trade-off  of baseline network, {bootstrap subsampling}, {adaptive residual learning}, and {iterative aggregation} using (a) HCP and (b) fastMRI dataset, respectively.}
 \label{table:no_param}
 \end{table}


As for the attention module, 
we employed MLP  for bootstrap subsampling.
Since the hidden unit of MLP was set to 64, the additional number of learnable parameters is $N \times 64 \times 2 + 64 +N$,
where $N$ is the number of intermediate neural network outputs from the bootstrap subsampled $k$-space data (recall that we used $N=10$ and 4 for HCP and 
fastMRI data set, respectively).
For the case of HCP data set, this leads to 1355 additional trainable parameters. This corresponds to
$5.4\times 10^{-3} \%$ complexity overhead, which can be considered negligible.
For other expressivity enhancement schemes, 1$\times$1 convolution layer was used.
In this case, the required additional parameter is $N\times (2N_c)^2 + 2N_c$, where
$N$ and $N_c$ denotes the number of intermediate results and coils, respectively (recall that we use $N=2$ and 4 for the case
of adaptive residual learning and iterative aggregation, respectively). 
For the HCP data set, this corresponds to 2080 and 4160 additional parameters for
adaptive residual learning and iterative aggregation, respectively. This again
leads to negligible complexity overhead of $8.3\times 10^{-3}\%$ and $16.4\times 10^{-3}\%$, respectively. 
Despite the negligible overhead, the performance enhancement was significant. For example,  the PSNR value using the {iterative aggregation} was 43.826 dB, compared to the baseline network with 39.675 dB as shown in Table \ref{table:no_param}.
The results for the fastMRI data in Table \ref{table:no_param} also confirmed the negligible complexity increase.

\begin{figure}[h] 	
\center{ 
\hspace*{-0.3cm}\includegraphics[width=7.5cm]{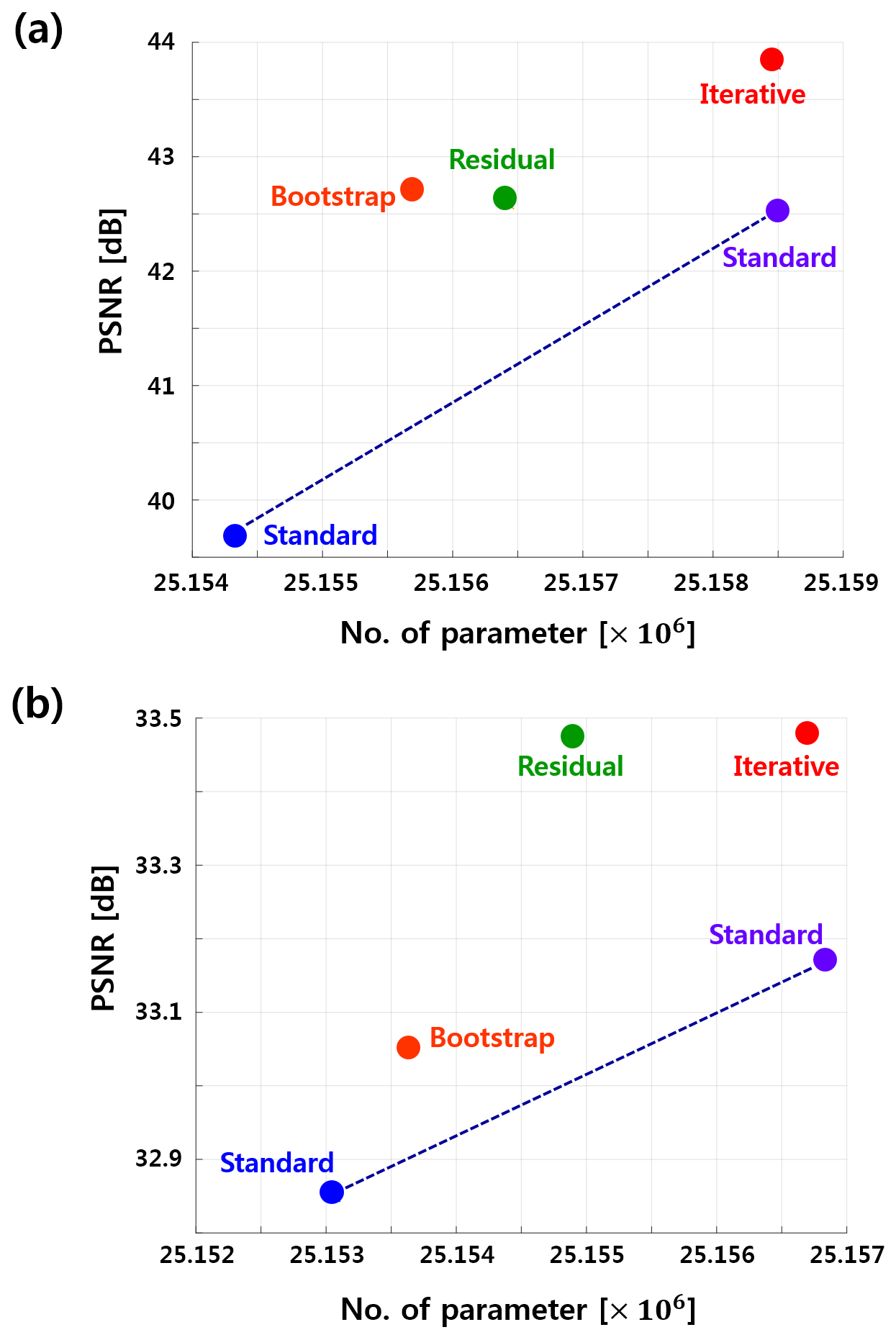}
}
\caption{The proposed schemes only slightly increases the number of parameters, improving the PSNR for the (a) HCP dataset at $R=15.42$ and (b) fastMRI dataset at $R=3.25$. Even the standard learning with more network capacity provided inferior performance than our schemes.}
\label{fig:param_psnr}
\end{figure}

We also compared the performance improvement using a baseline U-Net by simply changing the number of parameters.
Specifically, the number of channels was increased to make the total number of parameters for HCP dataset and fastMRI dataset up to 25,158,496 and 25,156,834, respectively. As shown in Fig. \ref{fig:param_psnr}, although the simple parameter increase improves PSNR, but
the performance enhancement was not remarkable compared to the
 proposed expressivity enhancement schemes. More specifically, {iterative aggregation} outperformed  U-Net with similar number of parameters
 up to about 1.30 dB and 0.31 dB in terms of PSNR for HCP dataset and fastMRI dataset, respectively.
This result confirmed that the performance improvement  of the proposed methods was not just due to the increase in complexity of the network, but from the synergistic expressivity enhancement that is not possible by simply increasing the number of channels. 

\subsection{Comparison with existing works}
\add{
Recursion have been employed to improve performance of the neural network in super-resolution\cite{kim2016deeply} and MR acceleration \cite{sun2018compressed} tasks. However, there is a significant difference between the proposed iterative aggregation scheme and previous works. Our study is the first to theoretically show why this iteration leads to improved performance of the networks, whereas previous studies have shown the effects only empirically based on experimental results.
In \cite{kim2016deeply}, although the weights $\wb = \{w_n\}_{n=1}^N$ for the aggregation of the intermediate results are learnable parameters that are adjusted during training, at inference stage these are held still.
In contrast, the attention module in the proposed scheme, which estimates each weight $w_n (z)$ based on each intermediate result $ \overbrace{\left(\Tc_{\Theta} \circ \cdots \circ \Tc_{\Theta}\right)}^{\text{$n$-times}} (\z)$, is adapted to the inputs not only at training stage but also at test time. 
The attention module is therefore effective to faithfully reflect the distribution of intermediate results even in the inference stage.
In \cite{sun2018compressed}, the network consists of $P$ blocks, and each block is recursively reused $N$ times. However, the final reconstruction image is the output of the last $P$-th block, and the intermediate results of the 
$P$ blocks are not aggregated for the final output \cite{sun2018compressed}. This is largely different from the proposed iterative aggregation scheme.}

\add{Furthermore, we investigated the role of ReLU. Our analysis shows that the input/feature space are divided into two disjoint regions by the hyperplane determined by each neuron thanks to the ReLU activation pattern. Although  ReLU  has been often viewed as a simple shrinkage operation, our perspective of deep neural network provides another closer link to the classical basis pursuit algorithm as explained in Section \ref{sec:CNN vs. Basis Pursuit}. More specifically, the neural network works as an effective multiplexer to different linear representations according to the ReLU activation pattern that is determined by the input. 
}

\section{Conclusion}
\label{conclusion}

In this paper, we investigated a systematic approach to improve the reconstruction performance of deep neural networks
for MRI reconstruction from accelerated acquisition.
We reviewed the recent geometric understanding of the deep convolutional neural network to show that the expressivity
and redundant representation are two important aspects of neural networks for image reconstruction.
Inspired by the fact that skipped connection achieves two goals,
  we proposed a systematic expressivity enhancement scheme that improves the expressivity of the underlying neural network with negligible complexity increase.
The performance improvement using the proposed expressivity improvement schemes were verified using various simulation and in vivo experiments. 
We believe that the proposed schemes can be applied not only to accelerated MRI but also to various inverse problems.

\add{One limitation of our current work is that our experimental validation of the proposed expressivity improvement schemes were only based on U-net. It is expected that the expressive power of other networks such as densely connected CNN (DenseCNN) \cite{huang2017densely}, which exploits a large number of skip connections, can be also analyzed
and further enhanced using the same proposed schemes.  
Another limitation of the current study is the lack of the analysis of batch normalization \cite{ioffe2015batch}.
 Batch normalization has been used frequently to reduce internal covariance shift which is the change in the distribution of network activations due to the change in network parameters during training. However, the theoretical analysis of batch normalization is an active area of research\cite{ba2016layer,bjorck2018understanding,ulyanov2016instance,salimans2016weight}, and there exists no standard theoretical analysis. 
 Therefore, we leave the investigation of these open problems for a possible future work of this study.}


\bibliographystyle{IEEEtran}

\end{document}